\documentclass[10pt,twocolumn,letterpaper]{article}

\usepackage{cvpr}
\usepackage{times}
\usepackage{epsfig}
\usepackage{graphicx}
\usepackage{amsmath}
\usepackage{amssymb}
\usepackage{pdfpages}

\usepackage[pagebackref=true,breaklinks=true,letterpaper=true,colorlinks,bookmarks=false]{hyperref}

\cvprfinalcopy % *** Uncomment this line for the final submission

\def\cvprPaperID{6644} % *** Enter the CVPR Paper ID here

\usepackage{wrapfig}

\usepackage{pbox}
\usepackage[utf8]{inputenc} % allow utf-8 input
\usepackage[T1]{fontenc}    % use 8-bit T1 fonts
\usepackage{url}            % simple URL typesetting
\usepackage{booktabs}       % professional-quality tables
\usepackage{amsfonts}       % blackboard math symbols
\usepackage{amssymb}
\usepackage{nicefrac}  
\usepackage{microtype}      % microtypography
\usepackage{graphicx}
\usepackage{subcaption}
\usepackage{multirow}
\usepackage{booktabs}
\usepackage{color}
\usepackage{bm}
\usepackage{soul}

\usepackage{algorithm}% http://ctan.org/pkg/algorithms
\usepackage[noend]{algpseudocode}
\usepackage{amsmath}
\newcommand*{\permcomb}[4][0mu]{{{}^{#3}\mkern#1#2_{#4}}}

\usepackage{amssymb}% http://ctan.org/pkg/amssymb
\usepackage{pifont}% http://ctan.org/pkg/pifont
\newcommand{\cmark}{\ding{51}}%
\newcommand{\xmark}{\ding{55}}%

\newcommand*\xoverline[2][0.75]{%
    \sbox{\myboxA}{$\m@th#2$}%
    \setbox\myboxB\null% Phantom box
    \ht\myboxB=\ht\myboxA%
    \dp\myboxB=\dp\myboxA%
    \wd\myboxB=#1\wd\myboxA% Scale phantom
    \sbox\myboxB{$\m@th\overline{\copy\myboxB}$}%  Overlined phantom
    \setlength\mylenA{\the\wd\myboxA}%   calc width diff
    \addtolength\mylenA{-\the\wd\myboxB}%
    \ifdim\wd\myboxB<\wd\myboxA%
       \rlap{\hskip 0.5\mylenA\usebox\myboxB}{\usebox\myboxA}%
    \else
        \hskip -0.5\mylenA\rlap{\usebox\myboxA}{\hskip 0.5\mylenA\usebox\myboxB}%
    \fi}

\usepackage[utf8]{inputenc} % allow utf-8 input
\usepackage[T1]{fontenc}    % use 8-bit T1 fonts
\usepackage{url}            % simple URL typesetting
\usepackage{booktabs}       % professional-quality tables
\usepackage{amsfonts}       % blackboard math symbols
\usepackage{amssymb}
\usepackage{nicefrac}       % compact symbols for 1/2, etc.
\usepackage{microtype}      % microtypography
\usepackage{graphicx}
\usepackage{multirow}
\usepackage{booktabs}
\usepackage{color}
\usepackage{bm}
\usepackage{soul}
\usepackage{enumitem}

\renewcommand{\arraystretch}{1.1}
\DeclareMathOperator*{\expectation}{\mathbb{E}}

\definecolor{ourdarkblue}{rgb}{0.0, 0.53, 0.74}
\definecolor{coolblack}{rgb}{0.0, 0.23, 0.64}

\ifcvprfinal\fi

\begin{document}

%%%%%%%%% TITLE
\title{Universal Source-Free Domain Adaptation}

\author{Jogendra Nath Kundu\thanks{Equal contribution.} \qquad Naveen Venkat\footnotemark[1] \qquad Rahul M V \qquad R. Venkatesh Babu\\
Video Analytics Lab, CDS, Indian Institute of Science, Bangalore\\
% {\tt\small jogendrak@iisc.ac.in, nav.naveenvenkat@gmail.com, rmvenkat@andrew.cmu.edu, venky@iisc.ac.in}
}

\maketitle
\thispagestyle{empty}

%%%%%%%%% ABSTRACT
\begin{abstract}
  There is a strong incentive to develop versatile learning techniques that can transfer the knowledge of class-separability from a labeled source domain to an unlabeled target domain in the presence of a domain-shift. Existing domain adaptation (DA) approaches are not equipped for practical DA scenarios as a result of their reliance on the knowledge of source-target label-set relationship (e.g. Closed-set, Open-set or Partial DA). Furthermore, almost all prior unsupervised DA works require coexistence of source and target samples even during deployment, making them unsuitable for real-time adaptation. Devoid of such impractical assumptions, we propose a novel two-stage learning process. 1) In the Procurement stage, we aim to equip the model for future source-free deployment, assuming no prior knowledge of the upcoming category-gap and domain-shift. To achieve this, we enhance the model’s ability to reject out-of-source distribution samples by leveraging the available source data, in a novel generative classifier framework. 2) In the Deployment stage, the goal is to design a unified adaptation algorithm capable of operating across a wide range of category-gaps, with no access to the previously seen source samples. To this end, in contrast to the usage of complex adversarial training regimes, we define a simple yet effective source-free adaptation objective by utilizing a novel instance-level weighting mechanism, named as Source Similarity Metric (SSM). A thorough evaluation shows the practical usability of the proposed learning framework with superior DA performance even over state-of-the-art source-dependent approaches. Our implementation is available on github\footnote{Code: \textbf{\url{https://github.com/val-iisc/usfda}}}.
\end{abstract}

\vspace{-2mm}
\section{Introduction}
% \vspace{-1mm}
\noindent Deep learning models have proven to be highly successful over a wide variety of tasks~\cite{krizhevsky2012imagenet,ren2015faster}. However, a majority of these are heavily dependent on access to a huge amount of labeled data to achieve a reliable level of generalization. A recognition model trained on a certain  distribution of labeled samples (source domain) %: $\{(x_s,y_s):x_s\sim p, y_s\in \mathcal{C}_s\}$) 
often fails to generalize~\cite{chen2017no} when deployed in a new environment (target domain) with discrepancy in the data distribution~\cite{shimodaira2000improving}. %: $\{x_t:x_t\sim q\}$). 
Unsupervised Domain Adaptation (DA) algorithms seek to minimize this discrepancy without accessing the target label information, either by learning a domain invariant feature representation~\cite{long2015learning,kumar2018co,ganin2016domain,tzeng2015simultaneous}, or by learning independent transformations~\cite{long2016unsupervised,nath2018adadepth} to a common latent representation through adversarial distribution matching~\cite{tzeng2017adversarial,kundu2019_um_adapt}.

\begin{figure}[!t]
    \centering
    \includegraphics[width=1.0\columnwidth]{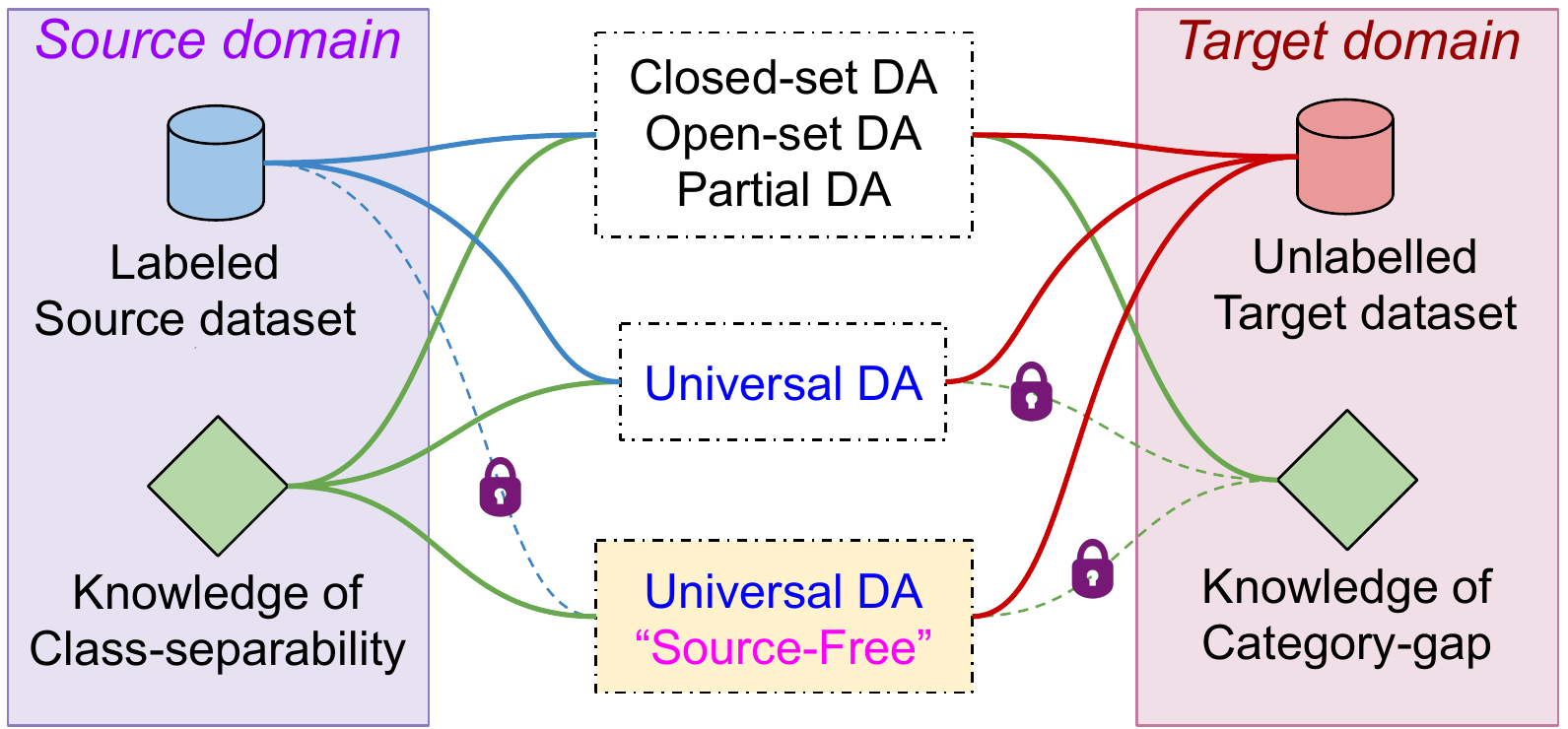}
    \vspace{-6mm}
    \caption{\small We address unsupervised domain adaptation in absence of source data (\textit{source-free}), without any category-gap knowledge (\textit{universal}). A lock indicates ``no access'' during adaptation.
    %Conceptual motivation: as a reliable negative sample.
    % For Ours and UDA~\cite{UDA_2019_CVPR}: Colored dotted lines indicate the various possible adaptation scenarios.
    }
    \vspace{-4mm}
    \label{fig_1}
\end{figure}
%%%%%%%%%%%%%%%%%%%%%%%%%%%%%%%%%%%%%%

\begin{figure}[!t]
    \centering
    \includegraphics[width=\linewidth]{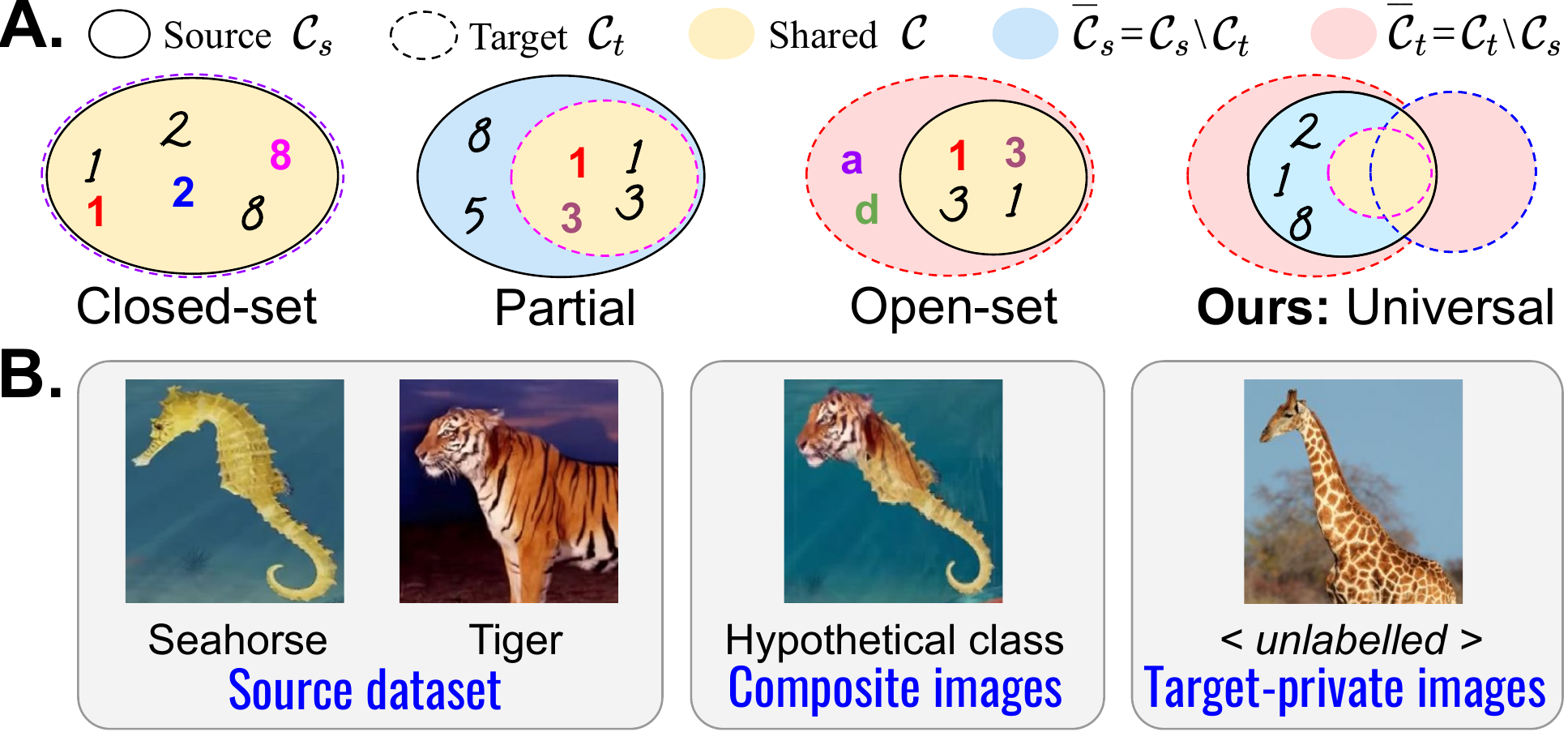}
    \vspace{-5mm}
    \caption{\small a) Various label-set relationships (\textit{category-gap}). b) Composite image as a reliable negative sample.}
    \label{fig:fig_2}
    \vspace{-4mm}
\end{figure}

Most of the existing approaches~\cite{saito2018maximum,zhang2018collaborative,tzeng2017adversarial} assume a shared label set between the source and the target domains (\ie $\mathcal{C}_s=\mathcal{C}_t$), \ie 
% which is often regarded as 
\textit{Closed-Set DA} (Fig.~\ref{fig:fig_2}{\color{red}A}). Though this assumption helps gain various insights for DA algorithms~\cite{ben2007analysis}, it rarely holds true in real-world scenarios. Recently, researchers have independently explored two broad adaptation settings by partly relaxing the \textit{Closed-Set} assumption (see Fig.~\ref{fig:fig_2}{\color{red}A}). In the first kind, \textit{Partial DA}~\cite{zhang2018importance,cao2018partial1,cao2018partial2}, the target label space is considered as a subset of the source label space (\ie $\mathcal{C}_t \subset \mathcal{C}_s$). This setting is more suited for large-scale universal source datasets, which will almost always subsume the label set of a wide range of target domains. However, the availability of such a large-scale source is highly questionable for a wide range of input domains.
In the second kind, \textit{Open-set DA}~\cite{saito2018open,OpenSetbaktashmotlagh2018learning,openSetge2017generative}, the target label space is considered as a superset of the source label space (\ie $\mathcal{C}_t \supset \mathcal{C}_s$). The major challenge in this setting is to detect target samples from the unobserved categories (similar to detection of out-of-distribution samples~\cite{malinin2018predictive}) in a fully-unsupervised scenario. Apart from the above two extremes, certain works define a partly mixed scenario by allowing a “private” label set for both source and target domains (\ie $\mathcal{C}_s \setminus \mathcal{C}_t \neq \emptyset$ and $\mathcal{C}_t \setminus \mathcal{C}_s \neq \emptyset $) but with extra supervision such as few-shot labeled data~\cite{luo2017label} or the knowledge of common categories~\cite{firstICCV}.

Most of the prior approaches~\cite{tzeng2017adversarial,saito2018open,cao2018partial1} consider each scenario in isolation and propose independent solutions. Thus, the knowledge of the relationship between the source and the target label space (category-gap) is required to carefully choose whether to apply \textit{Closed-set}, \textit{Open-set} or \textit{Partial} DA algorithm for the problem in hand. 
%Moreover, in the absence of prior knowledge regarding the relationship of label space, such approaches fail to offer a suitable direction. Motivated by this, we propose a unified solution for domain adaptation, without accessing the knowledge of the relationship among the source and target categories (\ie relationship between $\mathcal{C}_s$ and $\mathcal{C}_t$). 
Furthermore, all the prior unsupervised DA works require the coexistence of source and target samples even during deployment, hence are not \textit{source-free}. This is highly impractical, as labeled source data may not be accessible after deployment due to several reasons. Many datasets are withheld due to privacy concerns (\eg biometric data) \cite{lopes2017data_dfkd} or simply due to the proprietary nature of the dataset. Moreover, in real-time deployment scenarios \cite{wu2019distilledpersonreid}, training on the entire source data is not feasible due to computational limitations. 
Even otherwise, an accidental loss (\eg data corruption) of the source data renders the prior unsupervised DA methods non-viable for a future model adaptation  \cite{lwf}.
% Furthermore, all the prior unsupervised domain adaptation works operate under the assumption of availability of labeled source data~\cite{OpenSetbaktashmotlagh2018learning,zhang2018importance,cao2018partial2} during the adaptation process. This is highly impractical, as the labeled source data may not be accessible after deployment due to licensing or other copyright limitations. 
Acknowledging these issues, we aim to formalize a unified solution for unsupervised DA completely devoid of these limitations. Our problem setting is illustrated in Fig.~\ref{fig_1} (note \textit{source-free} and universal).

The available DA techniques heavily rely on the adversarial discriminative~\cite{tzeng2017adversarial,zhang2018collaborative,saito2018maximum} strategy. 
% Such approaches 
Thus, they require access to the source samples to reliably characterize the source domain distribution. Clearly, such approaches are not equipped to operate in a \textit{source-free} setting.
%. Such approaches require access to the source samples to reliably characterize the source domain distribution, which is utilized to minimize the target domain discrepancy. Clearly, such approaches are not equipped to operate in a source-free setting. 
Though a generative model can be used as a memory-network~\cite{sankaranarayanan2018generate,bousmalis2017unsupervised} to realize \textit{source-free} adaptation, such a solution is not scalable for large-scale source datasets (\eg ImageNet~\cite{imagenet}), as it introduces unnecessary additional parameters along with the associated training difficulties~\cite{salimans2016improved}. As a novel alternative, we hypothesize that, to facilitate \textit{source-free} adaptation, the source model should have the ability to reject samples that are out of the source data distribution~\cite{hendrycks2018deep}. 

% (\ie negative source samples). 

In general, fully-discriminative deep models have a tendency to over-generalize for regions not covered by the training set, hence are highly confident in their predictions even for negative samples~\cite{lee2018training}. Though this problem can be addressed by training the source model on a negative source dataset, a wrong choice of negative data makes the model incapable of rejecting unknown target samples encountered after deployment~\cite{Shafaei2019}.
%Therefore, we plan to leverage the commonness of semantic categories between the source and target domain as an important cue to synthetically simulate negative source samples. 
Aiming towards a data-free setting, we hypothesize that the target samples have similar local part-based features as found in the source data, which also holds for novel target categories as encountered in \textit{Open-set} DA. 
% Thus, the target-private categories could be simulated by combining local parts in source images. 
For example, consider an animal classification model (see Fig.~\ref{fig:fig_2}{\color{red}B}) where the deployed environment contains novel target categories unobserved in the source dataset (\eg Giraffe). Here, the composition of local regions (\eg body-parts) between pairs of source images drawn from different categories (\eg Seahorse and Tiger) can be used to synthetically generate hypothetical negative classes which can act as a proxy for the unobserved animal categories. Such synthetic samples are a better approximation of the expected characteristics (\eg long-neck) in the deployed target environment, as compared to samples from other unrelated datasets.

% For example, for a digit classification model (\eg MNIST), images with random strokes in varied colors are more suitable negative samples as compared to samples from Fashion-MNIST or CIFAR. Motivated by this we propose a systematic procedure to synthetically simulate labeled negative samples by compositing %local regions between pairs of positive source samples drawn from different source categories (see Fig.~\ref{fig_2}{B}). 

%incorporate a novel image composition approach to simulate negative samples by compositing image regions from different category to from the actual source data.
In summary, we propose a convenient DA framework, which is equipped to address Universal Source-Free Domain Adaptation. A thorough evaluation shows the practical usability of our approach with superior DA performance even over state-of-the-art source dependent approaches, across a variety of unknown label-set relationships.

%%%%%%%%%% figure 1 %%%%%%%%%%%%%%%%%%
\begin{figure*}[!ht]
    \centering
    \includegraphics[width=0.97\linewidth]{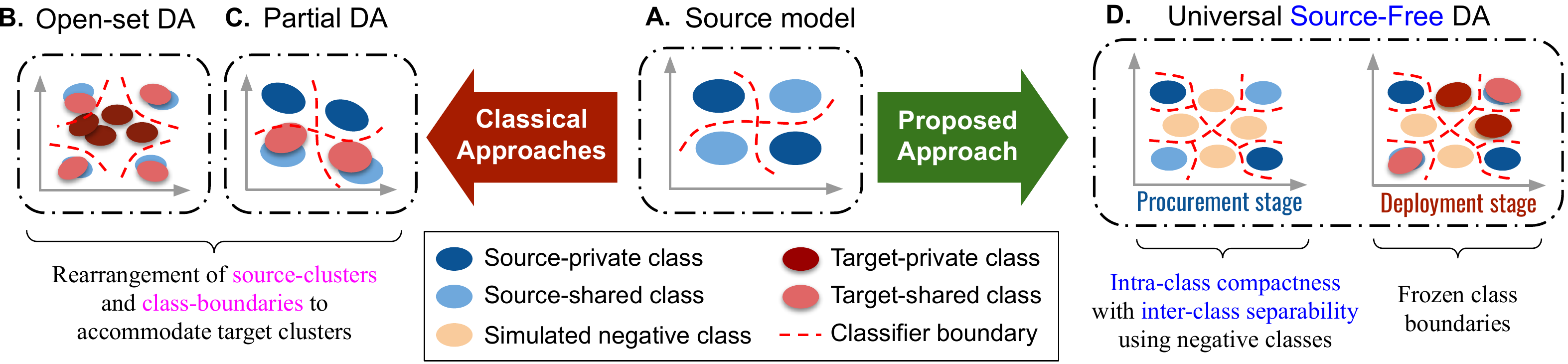}
    \vspace{-3mm}
    \caption{\small Latent space cluster arrangement during adaptation (see Section \ref{sec_procurement_stage}{\color{red}.1}). %B. Generated negative samples using randomly created spline segments (in purple), C. Proposed two-stage architecture.
    }
    \vspace{-3mm}
    \label{fig_cluster_rearrangement}
\end{figure*}
%%%%%%%%%%%%%%%%%%%%%%%%%%%%%%%%%%%%%%

\vspace{-0.7mm}
\section{Related work}
\vspace{-1.3mm}
We briefly review the available domain adaptation methods under the three major divisions according to the assumption on label-set relationship. 
\textbf{a) Closed-set DA.} The cluster of prior \textit{closed-set DA} works focuses on minimizing the domain gap at the latent space either by minimizing well-defined statistical distance functions~\cite{wang2014flexible,duan2012domain,zhang2013domain,office} or by formalizing it as an adversarial distribution matching problem~\cite{tzeng2017adversarial,kang2018deep,long2018conditional,hu2018duplex,hoffman2017cycada} inspired from the Generative Adversarial Nets~\cite{goodfellow2014generative}. %Besides the learning approach, architecture designs preferring either domain agnostic~\cite{ganin2016domain,tzeng2015simultaneous} or domain-specific features~\cite{tzeng2017adversarial,liu2016coupled,long2016unsupervised} has also been well-studied in previous works. 
Certain prior works~\cite{sankaranarayanan2018generate,zhu2017unpaired,hoffman2017cycada} use the GAN framework to explicitly generate target-like images translated from the source image samples, which is also regarded as pixel-level adaptation~\cite{bousmalis2017unsupervised} in contrast to other feature level adaptation works~\cite{nath2018adadepth,tzeng2017adversarial,long2015learning,long2016unsupervised}. 
\textbf{b) Partial DA.} \cite{cao2018partial1} proposed to achieve adversarial class-level matching by utilizing multiple domain discriminators furnishing a class-level and an instance-level weighting for individual data samples. \cite{zhang2018importance} proposed to utilize importance weights for source samples depending on their similarity to the target domain data using an auxilliary discriminator. To effectively address the problem of \textit{negative-transfer}~\cite{wang2018characterizing}, \cite{cao2018partial2} employed a single discriminator to achieve both adversarial adaptation and class-level weighting of source samples. 
\textbf{c) Open-set DA.} \cite{saito2018open} proposed a more general \textit{open-set DA} setting without accessing the knowledge of source-private labels in contrast to~\cite{panareda2017open}. They extended the classifier to accommodate an additional ``unknown” class, which is adversarially trained against other source classes to detect target-private samples. 
\textbf{d) Universal DA.} \cite{UDA_2019_CVPR} proposed the \textit{Universal DA} setting, which requires no prior knowledge of label-set relationship (see Fig.~\ref{fig:fig_2}{\color{red}A}), similar to our proposed setting, but considers access to both source and target samples during adaptation.

% \vspace{-0.5mm}
\section{Proposed approach}
% \vspace{-0.5mm}
%In this section, we first briefly introduced the notations and problem setting of universal domain adaptation. Subsequently, we present the proposed unified domain adaptation framework in a fully source-free setting.

%\subsection{Problem formulation}
Our approach to solve the \textit{source-free} domain adaptation problem is broadly divided into a two stage process. Note, \textit{source-free} DA means the adaptation step is \textit{source-free}. See Supplementary for a notation table.

\vspace{1mm}
\noindent \textbf{a) Procurement stage.} In this stage, we have a labeled source dataset, $\mathcal{D}_s=\{(x_s,y_s): x_s\sim p,\, y_s\in \mathcal{C}_s \}$, where $p$ is the distribution of source samples and $\mathcal{C}_s$ denotes the label-set of the source domain. Here, the prime objective is to equip the model for a future \textit{source-free} adaptation, 
where the model will encounter an unknown domain-shift and category-gap in the target domain.
% \ie the \textit{Deployment} stage, in the presence of a distribution discrepancy between the source and the target domains. 
% (i.e. \textit{domain-shift}).
%the source-free domain adaptation without accessing the unlabeled target domain data, $\mathcal{D}_t=\{(x_t): x_t\sim q \}$ with $q$ being the distribution of target samples. 
To achieve this we rely on an artificially generated negative dataset, $\mathcal{D}_n=\{(x_n,y_n):x_n\sim p_n,\,y_n\in\mathcal{C}_n\}$, where $p_n$ is the distribution of negative source samples such that $\mathcal{C}_n \cap \mathcal{C}_s = \emptyset$. %Here $\mathcal{C}_n$ is the label-set of negative source samples. 

%We train a model with samples from both $\mathcal{D}_n$ and $\mathcal{D}_s$ for a k-way classification, where $k=\vert\mathcal{C}_s\vert+\vert\mathcal{C}_n\vert$. 
%The training procedure and accessibility of a suitable $\mathcal{D}_n$ dataset is elaborated in Section XX.

\vspace{1mm}
\noindent \textbf{b) Deployment stage.} After obtaining a trained model from the \textit{Procurement} stage, the model will have its first encounter with the unlabeled target domain samples from the deployed environment. We denote the unlabeled target data by $\mathcal{D}_t=\{x_t: x_t\sim q \}$, where $q$ is the distribution of target samples. {Note that, the source dataset $\mathcal{D}_s$ from the \textit{Procurement} stage is inaccessible during adaptation in the \textit{Deployment} stage.} Suppose that, $\mathcal{C}_t$ is the label-set of the target domain. In the Universal setting~\cite{UDA_2019_CVPR}, we do not have any knowledge of the relationship between $\mathcal{C}_t$ and $\mathcal{C}_s$. Nevertheless, without the loss of generality,
we define the shared labels as $\mathcal{C} = \mathcal{C}_s\cap\mathcal{C}_t$ and the private label-set for the source and the target domains as $\overline{\mathcal{C}}_s = \mathcal{C}_s \setminus \mathcal{C}_t$ and $\overline{\mathcal{C}}_t = \mathcal{C}_t \setminus \mathcal{C}_s$ respectively. %\hl{About commonness}. 

%%%%%%%%%% figure 1 %%%%%%%%%%%%%%%%%%
%\begin{wrapfigure}{r}{0.71\textwidth}
\begin{figure*}[!ht]
    \centering
    \includegraphics[width=1\linewidth]{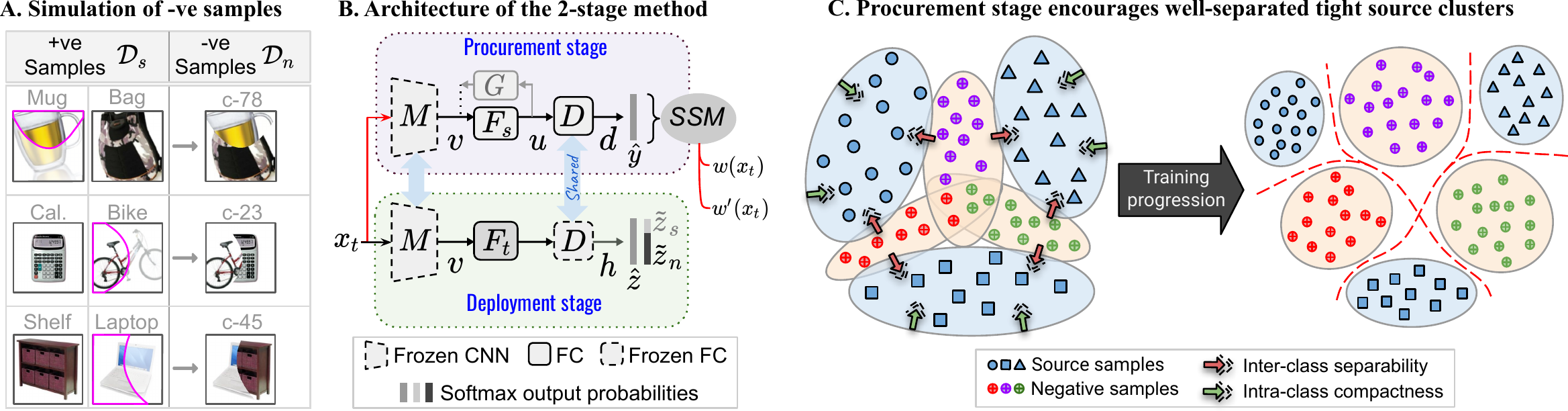}
    \vspace{-5mm}
    \caption{\small \textbf{A)} Simulated labeled negative samples using randomly created spline segments (in pink), \textbf{B)} Proposed architecture, \textbf{C)} \textit{Procurement} stage promotes intra-class compactness with inter-class separability.
    %Arrangement of positive latent clusters after procurement.
    }
    \label{fig_architecture} %\vspace{-1mm}
    \vspace{-3mm}
\end{figure*}
%\end{wrapfigure}
%%%%%%%%%%%%%%%%%%%%%%%%%%%%%%%%%%%%%%

\subsection{Learning in the Procurement stage}
\label{sec_procurement_stage}

% \vspace{-0.5mm}

\noindent \textbf{3.1.1. $\;$Challenges.} 
%%%% para
The available DA techniques heavily rely on the adversarial discriminative~\cite{tzeng2017adversarial,saito2018maximum} strategy. Thus, they require access to the source data to reliably characterize the source distribution. Further, these approaches are not equipped to operate in a \textit{source-free} setting.
%. Such approaches require access to the source samples to reliably characterize the source domain distribution, which is utilized to minimize the target domain discrepancy. Clearly, such approaches are not equipped to operate in a source-free setting. 
Though a generative model can be used as a memory-network~\cite{sankaranarayanan2018generate,bousmalis2017unsupervised} to realize \textit{source-free} adaptation, such a solution is not scalable for large-scale source datasets (\eg ImageNet~\cite{imagenet}), as it introduces unnecessary additional parameters alongside the associated training difficulties~\cite{salimans2016improved}. %As a novel alternative, we hypothesize that, to facilitate \textit{source-free} adaptation, the source model should have the ability to reject samples that are out of the source data distribution~\cite{hendrycks2018deep}.
This calls for a fresh analysis of the requirements beyond the existing solutions.
% found in literature.

In a general DA scenario, with access to source samples in the \textit{Deployment} stage (specifically for \textit{Open-set} or \textit{Partial} DA), a widely adopted approach is to learn domain invariant features. In such approaches, the placement of source category clusters is learned in the presence of unlabeled target samples which obliquely provides a supervision regarding the relationship between $\mathcal{C}_s$ and $\mathcal{C}_t$. For instance, in case of \textit{Open-set} DA, the source clusters may have to disperse to make space for the clusters from target-private $\overline{\mathcal{C}}_t$ (see Fig.~\ref{fig_cluster_rearrangement}{\color{red}A} to \ref{fig_cluster_rearrangement}{\color{red}B}). Similarly, in \textit{partial} DA, the source clusters may have to rearrange themselves to keep all the target shared clusters ($\mathcal{C} = \mathcal{C}_t$) separated from the source private $\overline{\mathcal{C}}_s$ (see Fig.~\ref{fig_cluster_rearrangement}{\color{red}A} to \ref{fig_cluster_rearrangement}{\color{red}C}). However in a completely \textit{source-free} framework, we do not have the liberty to leverage such information as source and target samples never coexist together during training. Motivated by the adversarial discriminative DA technique~\cite{tzeng2017adversarial}, we hypothesize that, inculcating the ability to reject samples that are out of the source data distribution can facilitate future \textit{source-free} domain alignment using this discriminatory knowledge.
%target-domain adaptation in an adversarial manner.
%As a novel alternative, we hypothesize that, to facilitate future source-free adaptation, the source model should have the ability to reject samples that are out of the source data distribution. 
Therefore, in the \textit{Procurement} stage the overarching objective is two-fold. 

\vspace{-2mm}
\begin{itemize}[leftmargin=6mm]

\item  Firstly, we must aim to learn a certain placement of source clusters best suited for all kinds of \textit{category-gap} scenarios acknowledging the fact that, a \textit{source-free} scenario does not allow us to modify the placement in the presence of target samples during adaptation (Fig.~\ref{fig_cluster_rearrangement}{\color{red}D}). 
\vspace{-1.5mm}

\item Secondly, the model must have the ability to reject out-of-distribution samples, which is an essential requirement for {unsupervised adaptation} under domain-shift.

\end{itemize}

\vspace{-1mm}
\noindent \textbf{3.1.2. $\;$Solution.} 
In the presence of source data, we aim to restrain the model's domain and category bias which is generally inculcated as a result of the over-confident supervised learning paradigms. To achieve this goal, we adopt two regularization strategies viz. 
i) utilization of a labeled simulated negative source dataset to generalize for the latent regions not covered by the given positive source samples (see Fig.~\ref{fig_architecture}{\color{red}C}) and ii) regularization via generative modeling.
% i) regularization via generative modeling and ii) utilization of a labeled simulated negative source dataset to generalize for the latent regions not covered by the given positive source samples (see Fig.~\ref{fig_architecture}{\color{red}C}).

%Keeping in mind the above objectives, the labeled negative dataset must
% \textbf{Characteristics of negative source dataset.}
\vspace{1mm}
\noindent \textbf{How to configure the negative source dataset?}\hspace{1mm}  
While configuring $\mathcal{D}_n$, the following key properties have to be met. Firstly, latent clusters formed by the negative categories must lie in-between the latent clusters of positive source categories to enable a higher degree of intra-class compactness with inter-class separability (Fig.~\ref{fig_architecture}{\color{red}C}). Secondly, the negative source samples must enrich the source domain distribution without forming a new domain by themselves. This %Note that, the second requirement 
rules out the use of Mixup~\cite{mixup} or adversarial noise ~\cite{vada_adversarial} as negative samples in this scenario.
% -- \hl{(nav: Can`t tell this if we are using manifold mixup for negative samples. Reviewer will be confused)}.
Thus, we propose the following method to synthesize the desired negative source dataset.

%\hl{Aiming towards} a data-free setting, We hypothesize that the target samples have similar local part-based features as found in the source data, which also holds for private target categories as encountered in \textit{Open-set} DA. For example, consider an animal classification model (see Fig.~\ref{fig_1}{B}) where the deployed environment contains some categories unobserved in source dataset (\eg Horse).  Here, composition of local regions (\eg body-parts) between pairs of positive source samples drawn from different source categories (\eg Seahorse and Tiger) can be used to synthetically generate hypothetical negative classes which can act as a proxy for the unobserved animal categories. Such synthetic samples are a better approximation of the expected samples in the deployed target environment, as compared to samples from other unrelated datasets.

%To achieve this, we propose a partially generative classifier model, by training Gaussian prior distributions on the labeled source samples $\mathcal{D}_s$. Additionally,  we also introduce a negative source dataset $\mathcal{D}_n$ to enrich the the model's ability to reject semantically related negative samples. 

%%%%%%%%%%%%%%%%%%%%% Algorithm 1: GAN_Tree Training Algorithm %%%%%%%%%%%%%%%%%%%%%
\begin{algorithm*}[!t]
%\algsetup{linenosize=\tiny}
\small %\vspace{-3mm} %\small, \footnotesize, \scriptsize, or \tiny
\caption{\small Training algorithm in the \textit{Procurement} stage}
\label{algo_procurement}%\vspace{-6mm}
\begin{algorithmic}[1]
\State \textbf{input:} $(x_s,y_s)\in\mathcal{D}_s$, $(x_n,y_n)\in\mathcal{D}_n$;
%: Labeled samples from  $\mathcal{D}_s$, and $\mathcal{D}_n$ respectively.\\ \hspace{8mm}
$\:\:\theta_{F_s}$, $\theta_{D}$, $\theta_{G}$: Parameters of $F_s$, $D$ and $G$ respectively.
\State \textbf{initialization:} pretrain $\{\theta_{F_s}, \theta_{D}\}$ using cross-entropy loss on $(x_s, y_s)$ followed by initialization of the sample mean $\mu_{c_i}$ and covariance $\Sigma_{c_i}$ (at $u$-space) of $F_s \circ M(x_s)$ for $x_s$ from class $c_i$; $i=1,2,...{\vert\mathcal{C}_s\vert}$
\For{${iter} < {MaxIter}$}
\State $v_s=M(x_s)$; $\;u_s=F_s(v_s)$; $\:\hat{v}_s=G(u_s)$; $\:u_r\sim \mathcal{N}(\mu_{c_i},\Sigma_{c_i})$ for $i=1,2,...{\vert\mathcal{C}_s\vert}$; $\:\hat{u}_r =F_s\circ G(u_r)$%; $a_n = F_s\circ M(x_n)$ ; $\hat{u}_n =F_s\circ G(a_n)$
\vspace{0.5mm}
\State $\hat{y}_s^{(k_s)}=\sigma^{(k_s)}(D\circ F_s\circ M(x_s))$, and  $\hat{y}_n^{(k_n)}=\sigma^{(k_n)}(D\circ F_s\circ M(x_n))$ where $k_s$, $k_n$ are the indices of ground-truth class $y_s$, $y_n$ 
% \Statex \hspace{4mm} of ground-truth label $y_s$ and $y_n$ respectively.
\vspace{0.5mm}
\State $\mathcal{L}_{CE}=-\log\hat{y}_s^{(k_s)}-\alpha\log\hat{y}_n^{(k_n)}$; $\:\mathcal{L}_v=\vert v_s-\hat{v}_s \vert$; $\:\mathcal{L}_u = \vert u_r-\hat{u}_r \vert $
\vspace{0.5mm}
\State $\mathcal{L}_p = -\log( \exp(P(u_s|c_{k_s}))/\sum_{i=1}^{\vert\mathcal{C}_s\vert}\exp(P(u_s|c_i)))$, $\:$where $P(u_s|c_i)=\mathcal{N}(u_s|\mu_{c_i},\Sigma_{c_i})$
\vspace{0.5mm}
%\State $\theta_{F_s}, \theta_{D} := \underset{\theta_{F_s},\theta_{D}}
   % {\textrm{argmin}}\hspace{1mm} -\log{\frac{exp(d^{(k)})}{\sum_{k=1}^{\vert\mathcal{C}_s\vert+\vert\mathcal{C}_n\vert}exp(d^{(k)}) }}$; where $d = D\circ F_s \circ M(x)$

%\State $\theta_{F_s}, \theta_{D} := \underset{\theta_{F_s},\theta_{D}}
   % {\textrm{argmin}}\hspace{1mm} \mathcal{L}_r$; where $\mathcal{L}_r=\vert v_s-\hat{v}_s \vert+\lambda_1\vert u_r-\hat{u}_r \vert - \lambda_2\vert a_n-\hat{u}_n \vert$
    
%\State $\theta_{F_s} := \underset{\theta_{F_s}}
    %{\textrm{argmin}}\hspace{1mm} \mathcal{L}_p$; where $\mathcal{L}_p = -\log( \exp(P(x_s|y_s=c_i))/\sum_{i=1}^{\vert\mathcal{C}_s\vert}\exp(P(x_s|y_s=c_i)))$

\State Update $\theta_{F_s}$, $\theta_{D}$, $\theta_{G}$ by minimizing $\mathcal{L}_{CE}$, $\mathcal{L}_{v}$, $\mathcal{L}_{u}$, and $\mathcal{L}_{p}$ alternatively using separate optimizers.
\vspace{0.5mm}
\If{$(iter$ $\mathbin{\%}$ $UpdateIter == 0)$}
\State Recompute the sample mean ($\mu_{c_i}$) and covariance ($\Sigma_{c_i}$) %of $u_s$ for each class $c_i$ 
of $F_s \circ M(x_s)$ for $x_s$ from class
$c_i$;
\Statex\hspace{9mm} 
% \newline \indent \indent 
$i=1,2...\vert\mathcal{C}_s \vert$ (For $\mathcal{D}_n^{(b)}$: generate fresh latent-simulated negative samples using the updated priors)
\EndIf

\EndFor
\end{algorithmic}
\end{algorithm*}
%%%%%%%%%%%%%%%%%%%%%%%%%%%%%%%%%%%%%%%%%

% \begin{figure}[!t]
%     \centering
%     \includegraphics[width=0.98\columnwidth]{figures/procurement_compactness_single.pdf}\vspace{-2mm}
%     \caption{\small Achieving intra-class compactness and inter-class separability using negative dataset $\mathcal{D}_n$.}
%     \label{procurement_compactness}
%     % \vspace{-3.5mm}
%     \vspace{-3mm}
% \end{figure}

% \vspace{-1mm}
\textbf{Image-composition.} One of the key characteristics shared between the samples from source and unknown target domain is the semantics of the local part-related features specifically for image-based object recognition tasks. Relying on this assumption, we propose a systematic procedure to simulate the samples of $\mathcal{D}_n$ by randomly compositing local regions between a pair of images drawn from the source dataset $\mathcal{D}_s$ (see Fig.~\ref{fig_architecture}{\color{red}A} and Suppl. Algo. {1}).
%the same or different source categories $\mathcal{C}_s$. 
These composite samples $x_n$ created on image pairs from different positive source classes are expected to 
%stimulate the activation of the two categories, thereby, hence lying
lie in-between the two source clusters in the latent space, thus introducing a combinatorial amount of new class labels \ie $\vert\mathcal{C}_n\vert = \permcomb{C}{\vert\mathcal{C}_s\vert}{2}$. %This clearly satisfies the aforementioned characteristics.

This approach is motivated from and conforms with the observation in the literature, that one can indeed generate semantics for new classes using the known classes \cite{lampert2009learning,vinyals2016matching}. Intuitively, from the perspective of combining features, when local parts from two different positive source classes are combined, the resulting image would tend to produce activations for both the classes (due to the presence of salient features from both classes). Thus, the sample would fall near the decision boundary in-between the two clusters in the latent space. Alternatively, from the perspective of discarding features, as we mask-out regions in a source image $x_s$ (Fig.~\ref{fig_architecture}), the activation in the corresponding class $y_s$ reduces. Thus, the model would be less confident for such samples, thereby emulating the characteristics of a negative class.

\vspace{1mm}
\noindent \textbf{Training procedure.} The generative source classifier is divided into three stages; i) backbone-model $M$, ii) feature extractor $F_s$, and iii) classifier $D$ (see Fig.~\ref{fig_architecture}{\color{red}B}). The output of the backbone-model is denoted as $v=M(x)$, where $x$ is drawn from either $\mathcal{D}_s$ or $\mathcal{D}_n$. Following this, the output of $F_s$ and $D$ are represented as $u$ and $d$ respectively.
%$a=F_s\circ M(x)$ and $d = D(z)$ respectively. 

$D$ outputs a $K$-dimensional logit vector denoted as $d=[d^{(k)}]$ for $k=1, 2, ..., K$, where $K=\vert\mathcal{C}_s\vert+\vert\mathcal{C}_n\vert$. The individual class probabilities, $\hat{y}^{(k)}$ are obtained by applying softmax over the logits \ie 
% $\hat{y}^{(k)}=\textit{exp}(d^{(k)})/\sum_{k=1}^{K}\textit{exp}(d^{(k)})=\sigma^{(k)}(D\circ F_s\circ M(x))$. 
$\hat{y}^{(k)}=\sigma^{(k)}(D\circ F_s\circ M(x))$, where $\circ$ denotes function composition, $\sigma$ denotes the softmax activation and the superscript $(k)$ denotes the class-index.
%We define two other class probabilities over the truncated logit vector separately for positive and negative source classes denoted as $\tilde{y}_s^k = \exp(d^k)/\sum_{k=1}^{\vert\mathcal{C}_s\vert}\exp(d^k)$ and  $\tilde{y}_n^k = \exp(d^k)/\sum_{k=\vert\mathcal{C}_s\vert+1}^{\vert\mathcal{C}_s\vert+\vert\mathcal{C}_n\vert}\exp(d^k)$. 

Additionally, we define priors only for the positive source classes, $P(u_s|c_i)=\mathcal{N}(u_s|\mu_{c_i},\Sigma_{c_i})$ (for $i=1, 2, ..., {\vert\mathcal{C}_s\vert}$) at the intermediate embedding $u_s=F_s\circ M(x_s)$. Here, the parameters of the normal distributions are computed during training as shown in line-10 of Algo.~\ref{algo_procurement}. A cross-entropy loss over these prior distributions is defined as $\mathcal{L}_p$ (line-7 in Algo.~\ref{algo_procurement}), that effectively enforces intra-class compactness with inter-class separability (Fig.~\ref{fig_architecture}{\color{red}C}). 

Motivated by generative variational auto-encoder (VAE) setup~\cite{kingma2013auto}, 
we introduce a decoder $G$, which minimizes the cyclic reconstruction loss selectively for the samples $v_s$ from positive source categories and randomly drawn samples $u_r$ from the corresponding class priors (\ie losses $\mathcal{L}_v$ and $\mathcal{L}_u$ in line-6 of Algo.~\ref{algo_procurement}). This, along with a lower weightage $\alpha$ for the negative source categories (\ie at the cross-entropy loss $\mathcal{L}_{CE}$ in line-6 of Algo.~\ref{algo_procurement}), is incorporated to deliberately bias $F_s$ towards the positive source samples, considering the level of unreliability of the generated negative dataset.

%to account for the class imbalance
% This is done also keeping in mind the imbalance $\vert \mathcal{C}_s\vert \ll \vert \mathcal{C}_n\vert$.

%The placement of negative classes in-between other positive source clusters along with the partial generative framework not only avoids over-generalization to positive source samples at the embedding space $u$, but also enforces learning of compact boundaries which helps us achieve the best possible placement of source category clusters in the absence of target domain samples. See Fig.~\ref{procurement_compactness}{\color{red}}.

%. Here, $\hat{v}_s=G(u_s)$, $u_r\sim \mathcal{N}(u_s|\mu_{c_i},\Sigma_{c_i})$ for $i=1,2,...c_{\vert\mathcal{C}_s\vert}$, and $\hat{u} =F_s\circ G(u_r)$. 
%only for the samples from positive source categories \ie $v_s=M(x_s)$.
%Along with the explicitly constructed labeled negative samples $x_n$, we adopted a partially generative framework to improve the model's ability to reject out-of-distribution samples. The cross-entropy on prior probabilities enforces intra-class compactness with inter-class separability. The cluster of negative classes helps us 
%\textit{Typically, a generative classifier produces predictions by comparing between the likelihood of the labels for a given input, which is closely related to the “distance” of the input to the data manifold associated with a class.}

%[intra] This is further enriched 
%by a regularization, which aim to maximize entropy class probabilities
%by introducing a regularization targeting samples around the class boundaries \ie samples belonging to $c_{intra}$. 

\vspace{-1mm}
\subsection{Learning in the Deployment stage}
\vspace{-1mm}

\noindent \textbf{3.2.1. $\;$Challenges.}
We hypothesize that, the large number of negative source categories along with the positive source classes \ie $\mathcal{C}_s\cup\mathcal{C}_n$ can be interpreted as a universal source dataset, which can subsume label-set $\mathcal{C}_t$ of a wide range of target domains. % (\ie the \textit{Partial} DA setting). However, the available \textit{Partial} DA approaches~\cite{cao2018partial1,cao2018partial2} rely on
Moreover, we seek to realize a unified adaptation algorithm, which can work for a wide range of \textit{category-gaps}.
%Besides this, one of the major challenges is to come up with a unified algorithm, which can work for varied types of unknown category gap (\ie relationship between $\mathcal{C}_s$ and $\mathcal{C}_t$). 
However, a forceful adaptation of target samples to positive source categories will cause target-private samples to be classified as an instance of the source private or the common label-set, instead of being classified as "\textit{unknown}", i.e. one of the negative categories in $\mathcal{C}_n$. %\hl{Real-time: training a domain discriminator on the entire source data during adaptation is not feasible, hence design SSM for instance weighting in a "source-free" setting}

%Whereas such samples should be classified as "\textit{unknown}" or one of the negative categories in $\mathcal{C}_n$. 

%Thus, the proposed learning algorithm should be capable of distinguishing target samples from the common label space $\mathcal{C}$ in a fully unsupervised manner.

\vspace{1mm}
\noindent \textbf{3.2.2. $\;$Solution.} 
In contrast to domain agnostic architectures~\cite{UDA_2019_CVPR,cao2018partial1,saito2018maximum}, we resort to an architecture supporting domain specific features~\cite{tzeng2017adversarial}, as we must avoid disturbing the placement of source clusters obtained from the \textit{Procurement} stage.
This is an essential requirement to retain the task-dependent knowledge gathered from the source dataset.
%as we are restricted to disturb the placement of source clusters obtained from the aforementioned procurement stage.
Thus, we introduce a domain specific feature extractor denoted as $F_t$, whose parameters are initialized from the fully trained $F_s$ (see Fig.~\ref{fig_architecture}{\color{red}B}). 
%In contrast to prior approaches~\cite{UDA_2019_CVPR,cao2018partial2}, we do not introduce other ad-hoc network (such as critic or discriminator) in the proposed adaptation framework. 
Further, we aim to exploit the learned generative classifier from the \textit{Procurement} stage to complement for the purpose of separate ad-hoc networks (critic or discriminator) as utilized by the prior works~\cite{UDA_2019_CVPR,cao2018partial2}.
%We define two separate class probabilities over the truncated logit vector $d$ separately for positive and negative source classes denoted as $\tilde{y}_s^{(i)} = \exp(d^{(i)})/\sum_{i=1}^{\vert\mathcal{C}_s\vert}\exp(d^{(k)})$ and  $\tilde{y}_n^{(j)} = \exp(d^{(j)})/\sum_{k=\vert\mathcal{C}_s\vert+1}^{\vert\mathcal{C}_s\vert+\vert\mathcal{C}_n\vert}\exp(d^{(j)})$ respectively. Note that, here $d=D\circ F_s \circ M(x_t)$ \ie forwarding the target samples through the source feature extractor. 
%Similarly 

\vspace{1mm}
\textbf{a) Source Similarity Metric (SSM).} 
% We define a weighting factor (SSM) for each target sample $x_t$, as $w(x_t)$. A higher value of this metric indicates $x_t$'s similarity towards the positive source categories, specifically inclined towards the common label space $\mathcal{C}$.
For each target sample $x_t$, we define a weighting factor $w(x_t)$ called the SSM. A higher value of this metric indicates $x_t$'s similarity towards the positive source categories, specifically inclined towards the common label space $\mathcal{C}$.
%. Among all the positive source categories $\mathcal{C}_s$, such samples will be inclined towards the categories in the common label space $\mathcal{C}$. %However, we do not aim to achieve a hard distinction between unknown source private $\overline{\mathcal{C}}_s$ and the common label-set $\mathcal{C}$. 
Similarly, a lower value of this metric indicates $x_t$'s similarity towards the negative source categories $\mathcal{C}_n$, showing its inclination towards the private target labels $\overline{\mathcal{C}}_t$. Let, $p_{\bar s}$, $q_{\bar t}$ be the distribution of source and target samples with labels in $\overline{\mathcal{C}}_s$ and $\overline{\mathcal{C}}_t$ respectively. We define, $p_c$ and $q_c$ to denote the distribution of samples from source and target domains belonging to the shared label-set $\mathcal{C}$.  Then, the \textit{SSM} for the positive and negative source samples should lie on the two extremes, forming the inequality:

% \vspace{-8mm}
\vspace{-3mm}
\begin{small}
\begin{equation}
\label{eqn_1}
        % \setlength{\mathindent}{0pt}
        % \vspace{-3mm}
        \expectation_{x\sim p_n} \hspace{-4pt} w(x)  \approx \hspace{-3pt} \expectation_{x\sim q_{\bar{t}}} \hspace{-4pt} w(x) \hspace{0pt} < \hspace{-2pt} \expectation_{x\sim q_{c}} \hspace{-4pt} w(x) \hspace{0pt} < \hspace{-2pt} \expectation_{x\sim p_{c}} \hspace{-4pt} w(x)  \approx \hspace{-3pt} \expectation_{x\sim p_{\bar{s}}} \hspace{-4pt} w(x)
\end{equation}
\end{small}
\vspace{-3mm}
% \vspace{-2mm}

% $$
% \mathbb{E}_{x_n\sim p_n}w(x_n)\approx\mathbb{E}_{x_t\sim q_{\bar{t}}}w(x_t)\ <
% \mathbb{E}_{x_t\sim q_{c}}w(x_t) < \mathbb{E}_{x_s\sim p_{c}}w(x_s) \approx\mathbb{E}_{x_s\sim p_{\bar{s}}}w(x_s)
% $$
%Now, we need to construct the \textit{SSM} criterion for individual target samples according to the desired characteristics discussed above. 
To formalize the \textit{SSM} criterion we rely on the class probabilities defined at the output of source model only for the positive class labels, \ie $\hat{y}^{(k)}$ for $k=1,2...\vert\mathcal{C}_s\vert$. Note that, $\hat{y}^{(k)}$ is obtained by performing softmax over $\vert\mathcal{C}_s\vert+\vert\mathcal{C}_n\vert$ categories as discussed in the \textit{Procurement} stage.
%, \ie $\hat{y}^{(k)}=\textit{exp}(d^{(k)})/\sum_{k=1}^{K}\textit{exp}(d^{(k)})$ where $d=D\circ F_s \circ M(x_t)$. 
Finally, the \textit{SSM} $w$ and its complement $w^{\prime}$ are {defined as}, 
% \vspace{-3mm}

\vspace{-6mm}
\begin{small}
\begin{equation}\label{eqn_2}
\begin{aligned}
w(x_t) & = \max_{i=1...\vert\mathcal{C}_s\vert} {\exp(\hat{y}^{(i)})} \\
w^{\prime}(x_t) & = \max_{i=1...\vert\mathcal{C}_s\vert} {\exp(1-\hat{y}^{(i)})}
% \vspace{-1mm}
\end{aligned}
\end{equation}
\end{small}
% \vspace{-2mm}

% $$
% w(x_t) = \max_{i=1,2...\vert\mathcal{C}_s\vert}{\exp(\hat{y}^{(k)})}, \;\: and \;\:
% w^\prime(x_t) = \max_{i=1,2...\vert\mathcal{C}_s\vert}{\exp(1-\hat{y}^{(k)})}
% $$

We hypothesize that this definition will satisfy Eq.~\ref{eqn_1}, 
as a result of the generative learning strategy adopted in the \textit{Procurement} stage.
In Eq.~\ref{eqn_2} the exponent is used to further amplify separation between target samples from the shared label-set $\mathcal{C}$ and those from the private label-set $\overline{\mathcal{C}}_t$ ({Fig.~\ref{fig_ssm_sensitivity}{\color{red}A}}).
% In Eq.~\ref{eqn_2} the exponent is used to further amplify separation among the target samples from the shared $\mathcal{C}$ and private $\overline{\mathcal{C}}_t$ label-set ({see Fig.~\ref{fig_4}{\color{red}A}}).

%$\tilde{y}_s^{(i)} = \exp(d^{(i)})/\sum_{i=1}^{\vert\mathcal{C}_s\vert}\exp(d^{(k)})$ and  $\tilde{y}_n^{(j)} = \exp(d^{(j)})/\sum_{k=\vert\mathcal{C}_s\vert+1}^{\vert\mathcal{C}_s\vert+\vert\mathcal{C}_n\vert}\exp(d^{(j)})$ respectively. Note that, here $d=D\circ F_s \circ M(x_t)$ \ie forwarding the target samples through the source feature extractor $F_s$. 

\vspace{1mm}
\textbf{b)$\;$Source-free domain adaptation.$\;$} 
To perform domain adaptation, the objective function aims to move the target samples with higher \textit{SSM} value towards the clusters of positive source categories and vice-versa at the frozen source embedding, $u$-space (from the \textit{Procurement} stage). To achieve this, parameters of only $F_t$ network are allowed to be trained in the \textit{Deployment} stage. However, the decision of weighting the loss on target samples towards the positive or negative source clusters is computed using the source feature extractor $F_s$ \ie the \textit{SSM} in Eq.~\ref{eqn_2}. We define, the deployment model as $h=D\circ F_t \circ M(x_t)$ using the target feature extractor, with softmax predictions over $K$
%$\vert\mathcal{C}_s\vert+\vert\mathcal{C}_n\vert$ 
categories obtained as %$\hat{z}^{(k)}=\textit{exp}(h^{(k)})/\sum_{k=1}^{K}\textit{exp}(h^{(k)})$. 
$\hat{z}^{(k)}=\sigma^{(k)}(h)$. 
Thus, the primary loss function for adaptation is defined as,

% \vspace{-2mm}
\begin{equation}\label{eqn_3}
\begin{small}
    \begin{aligned}
    \mathcal{L}_{d1} = &~w(x_t) \cdot \Big( - \log({\textstyle \sum_{k=1}^{\vert\mathcal{C}_s\vert}\hat{z}^{(k)}}) \Big) ~~+ \\
    &~w^{\prime}(x_t) \cdot \Big( - \log({\textstyle \sum_{k=\vert\mathcal{C}_s\vert+1}^{\vert\mathcal{C}_s\vert+\vert\mathcal{C}_n\vert}\hat{z}^{(k)}}) \Big)
    \end{aligned}
    \end{small}
\end{equation}
% \vspace{-2mm}

% $$
% \mathcal{L}_{d1} = -w(x_t)\log({\textstyle \sum_{k=1}^{\vert\mathcal{C}_s\vert}\hat{z}^{(k)}}) -w^\prime(x_t)\log({\textstyle \sum_{k=1+\vert\mathcal{C}_s\vert}^{\vert\mathcal{C}_s\vert+\vert\mathcal{C}_n\vert}\hat{z}^{(k)}})
% $$

Additionally, in the absence of label information, there would be uncertainty in the predictions $\hat{z}^{(k)}$ as a result of distributed class probabilities. This leads to a higher entropy for such samples. Entropy minimization~\cite{grandvalet2005semi,long2016unsupervised} is adopted in such scenarios to move the target samples close to the highly confident regions (\ie positive and negative cluster centers from the \textit{Procurement} stage) of the classifier's feature space. % in the frozen source embedding space $u$. 
However, it has to be done separately for positive and negative source categories based on the \textit{SSM} values of individual target samples to effectively 
distinguish the target-private set from the full target dataset. %~\cite{saito2018open}. 
To achieve this, we define two different class probability vectors separately for the positive and negative source classes (Fig.~\ref{fig_architecture}{\color{red}B}) as, 

\vspace{-4mm}
\begin{small}
\begin{equation}
    \begin{aligned}
    \tilde{z}_s^{(i)} = \frac{\exp(h^{(i)})}{{\textstyle\sum_{j=1}^{\vert\mathcal{C}_s\vert}}\exp(h^{(j)})} ~~~~;~~~
    \tilde{z}_n^{(i)} =  \frac{\exp(h^{(i+\vert\mathcal{C}_s\vert)})}{{\textstyle\sum_{j=1}^{\vert\mathcal{C}_n\vert}}\exp(h^{(j+\vert\mathcal{C}_s\vert)})}
    \end{aligned}
\end{equation}
\end{small}
\vspace{-3mm}

% $\tilde{z}_s^{(i)} = \exp(h^{(i)})/{\textstyle\sum_{j=1}^{\vert\mathcal{C}_s\vert}}\exp(h^{(j)})$ and  $\tilde{z}_n^{(i)} = \exp(h^{(i+\vert\mathcal{C}_s\vert)})/{\textstyle\sum_{j=1}^{\vert\mathcal{C}_n\vert}}\exp(h^{(j+\vert\mathcal{C}_s\vert)})$ respectively (see Fig.~\ref{fig_architecture}{\color{red}B}).
% \begin{equation}\label{eqn_4}
% \begin{small}
%     \begin{aligned}
%     \tilde{z}_s^{(i)} = \exp(e^{(i)})/{\textstyle\sum_{j=1}^{\vert\mathcal{C}_s\vert}}\exp(e^{(j)}) \:\;\;\;\;  \tilde{z}_n^{(i)} = \exp(e^{(i+\vert\mathcal{C}_s\vert)})/{\textstyle\sum_{j=1}^{\vert\mathcal{C}_n\vert}}\exp(e^{(j+\vert\mathcal{C}_s\vert)})
%     \end{aligned}
%     \end{small}
% \end{equation}
%$\tilde{z}_s^{(i)} = \exp(e^{(i)})/\sum_{i=1}^{\vert\mathcal{C}_s\vert}\exp(e^{(i)})$ and  $\tilde{z}_n^{(j)} = \exp(e^{(j+\vert\mathcal{C}_s\vert)})/\sum_{j=1}^{\vert\mathcal{C}_n\vert}\exp(e^{(j+\vert\mathcal{C}_s\vert)})$ respectively. 
%Following this, 
We obtain the entropy of the target samples for the positive source classes as $H_s(x_t) =- \sum_{i=1}^{\vert\mathcal{C}_s\vert}\tilde{z}_s^{(i)}\log\tilde{z}_s^{(i)}$ and for the negative classes as $H_n(x_t) = -\sum_{i=1}^{\vert\mathcal{C}_n\vert}\tilde{z}_n^{(i)}\log\tilde{z}_n^{(i)}$.
% \vspace{-3mm}
% \begin{equation}\label{eqn_5}
% \begin{small}
%     \begin{aligned}
%     H_s(x_t) &=- \sum_{i=1}^{\vert\mathcal{C}_s\vert}\tilde{z}_s^{(i)}\log\tilde{z}_s^{(i)} \\
%     H_n(x_t) &= -\sum_{i=1}^{\vert\mathcal{C}_n\vert}\tilde{z}_n^{(i)}\log\tilde{z}_n^{(i)}
%     \end{aligned}
%     \end{small}
% \end{equation}
% \vspace{-4mm}
%  and  respectively. 
Subsequently, the entropy minimization is formulated as, 

\vspace{-3mm}
\begin{equation}\label{eqn_5}
\begin{small}
    \begin{aligned}
    \mathcal{L}_{d2} = w(x_t) \cdot H_s(x_t) + w^{\prime}(x_t) \cdot H_n(x_t)
    \end{aligned}
    \end{small}
\end{equation}
\vspace{-4mm}

% $$
% \mathcal{L}_{d2} = w(x_t)H_s(x_t)+w^\prime(x_t)H_n(x_t)
% $$
Thus, the final loss function for adaptation is $\mathcal{L}_d = \mathcal{L}_{d1}+\beta\mathcal{L}_{d2}$. Here $\beta$ is a hyper-parameter controlling the importance of entropy minimization during adaptation.
% Thus, the final loss function for adapting the parameters of $F_t$ is presented as $\mathcal{L}_d = \mathcal{L}_{d1}+\beta\mathcal{L}_{d2}$. Here $\beta$ is a hyper-parameter controlling the importance of entropy minimization during adaptation.

%\textbf{Inference}

%%%%%%%%%%%%%%%%%%%%%%%%%%%%%%%%%%%%%%%%%%%%%%%%%%%%%%%%%%%
%%%%%%%%%%%%%%% Experiments %%%%%%%%%%%%%%%%%%%%%%%%%%%%%%
%%%%%%%%%%%%%%%%%%%%%%%%%%%%%%%%%%%%%%%%%%%%%%%%%%%%%%%%%%%
%\input{main_naveen}
\section{Experiments}
\vspace{-1mm}
\noindent We perform a thorough evaluation of the proposed universal \textit{source-free} domain adaptation framework against prior state-of-the-art methods across multiple datasets. We also provide a comprehensive ablation study to establish generalizability of the approach across a variety of label-set relationships and justification of the various model components.

%\textbf{4.1\; Experimental Setup}
% \vspace{-1mm}
\subsection{Experimental Setup}
\vspace{-1mm}

\textbf{a) Datasets.} 
We resort to the experimental settings followed by \cite{UDA_2019_CVPR} ({UAN}). \textbf{Office-Home}~\cite{venkateswara2017deep} dataset consists of images from 4 different domains - Artistic (\textbf{Ar}), Clip-art (\textbf{Cl}), Product (\textbf{Pr}) and Real-world (\textbf{Rw}). 
% Alphabetically, the first 10 classes are selected as $\mathcal{C}$, the next 5 classes as $\overline{\mathcal{C}}_s$, and the rest {50} as $\overline{\mathcal{C}}_t$. 
\textbf{VisDA2017}~\cite{visda} dataset comprises of 12 categories with synthetic (\textbf{S}) and real (\textbf{R}) domains.
\textbf{Office-31} ~\cite{office} dataset contains images from 3 distinct domains - Amazon (\textbf{A}), DSLR (\textbf{D}) and Webcam (\textbf{W}). 
% We use the 10 classes shared by Office-31 and Caltech-256~\cite{gong2012geodesic} to construct the shared label-set $\mathcal{C}$ and alphabetically select the next 10 as $\overline{\mathcal{C}}_s$, with the rest 11 classes as $\overline{\mathcal{C}}_t$. 
To evaluate scalability, we use \textbf{ImageNet-Caltech} with 84 common classes (following \cite{UDA_2019_CVPR}).
% For all the datasets, we resort to the experimental settings followed by \cite{UDA_2019_CVPR} ({UAN}). \textbf{Office-Home}~\cite{venkateswara2017deep} dataset consists of images from 4 different domains - Artistic (\textbf{Ar}), Clip-art (\textbf{Cl}), Product (\textbf{Pr}) and Real-world (\textbf{Rw}). Alphabetically, the first 10 classes are selected as $\mathcal{C}$, the next 5 classes as $\overline{\mathcal{C}}_s$, and the rest {50} as $\overline{\mathcal{C}}_t$. \textbf{VisDA2017}~\cite{visda} dataset comprises of 12 categories with synthetic images as the source domain and natural images as the target domain, out of which, the first 6 are chosen as $\mathcal{C}$, the next 3 as $\overline{\mathcal{C}}_s$ and the rest as $\overline{\mathcal{C}}_t$. \textbf{Office-31} ~\cite{office} dataset contains images from 3 distinct domains - Amazon (\textbf{A}), DSLR (\textbf{D}) and Webcam (\textbf{W}). We use the 10 classes shared by Office-31 and Caltech-256~\cite{gong2012geodesic} to construct the shared label-set $\mathcal{C}$ and alphabetically select the next 10 as $\overline{\mathcal{C}}_s$, with the rest 11 classes as $\overline{\mathcal{C}}_t$. To evaluate scalability, \textbf{ImageNet-Caltech} is also considered with 84 common classes (following \cite{UDA_2019_CVPR}).
% inline with the setting in~\cite{UDA_2019_CVPR}. % cao2018partial1 to evaluate scalability of the approach.

\vspace{1mm}
\textbf{b) Simulation of labeled negative samples.} To simulate negative samples for training in the \textit{Procurement} stage, %we first augment the positive source images with random rotation, translation and color jittering. Following this, 
we first sample a pair of images, each from different categories of $\mathcal{C}_s$, to create unique negative classes in $\mathcal{C}_n$. Note that, we impose no restriction on how the hypothetical classes are created (\eg one can composite non-animal with animal). A random mask is defined which splits the images into two complementary regions using a quadratic spline passing through a central image region (see Suppl. Algo. {\color{red}1}). Then, the negative image is created by merging alternate mask regions as shown in Fig.~\ref{fig:fig_2}{\color{red}A}. For the {\textbf{I}}$\rightarrow${\textbf{C}} task of ImageNet-Caltech, the source domain ImageNet (\textbf{I}), having 1000 classes, results in a large number of possible negative classes (\ie $\vert\mathcal{C}_n\vert =  \permcomb{C}{\vert\mathcal{C}_s\vert}{2}$). We address this by randomly selecting only 600 of these negative classes for ImageNet (\textbf{I}), and 200 negative classes for Caltech ({\textbf{C}}) in the task {\textbf{C}}$\rightarrow${\textbf{I}}. 
% In a similar fashion, we generate latent-simulated negative samples only for the selected negative classes in these datasets. We compare the two models with different \textit{Procurement} stage training - (i) \textbf{\textit{USFDA-a}}: using image-composition as negative dataset, and (ii) \textbf{\textit{USFDA-b}}: using latent-simulated negative samples as the negative dataset. We use {\textit{USFDA-a}} for most of our ablations unless mentioned explicitly.

%(\textbf{\textit{USFDA-b}}).
% We denote the model trained with image-composition during \textit{Procurement} as \textbf{\textit{USFDA-a}}, and the one trained with latent-simulated negative samples as \textbf{\textit{USFDA-b}}.

%%%%%%%%%%%%%%%%%%%%%%% Table-1 Office to home %%%%%%%%%%%%%%%%%%%%%%%%%%

\begin{table*}[!t]
    \addtolength{\tabcolsep}{0pt}
    \centering
    \caption{\small Average per-class accuracy ($\mathcal{T}_{avg}$) for universal-DA tasks on \textbf{Office-Home} dataset (with $\vert{\mathcal{C}}\vert / \vert{\mathcal{C}_s\cup\mathcal{C}_t\vert}=0.15$). Scores for the prior works are directly taken from UAN~\cite{UDA_2019_CVPR}. Here, SF denotes support for \textit{source-free} adaptation.
    \vspace{-3mm}}
    \label{tab:table_1}
    \resizebox{1\textwidth}{!}{%
    \begin{tabular}{c|c|ccccccccccccc}
        \toprule
        \multirow{2}{30pt}{\centering Method} & \multirow{2}{*}{\centering SF} & \multicolumn{13}{c}{Office-Home} \\
        \cmidrule(lr){3-15}
        & & {Ar}$\rightarrow${Cl} & {Ar}$\rightarrow${Pr} & {Ar}$\rightarrow${Rw} & {Cl}$\rightarrow${Ar} & {Cl}$\rightarrow${Pr} & {Cl}$\rightarrow${Rw} & {Pr}$\rightarrow${Ar} & {Pr}$\rightarrow${Cl} & {Pr}$\rightarrow${Rw} & {Rw}$\rightarrow${Ar} & {Rw}$\rightarrow${Cl} & {Rw}$\rightarrow${Pr} & Avg \\
        \hline\hline
        ResNet~\cite{he2016deep} & \xmark & 59.37 & 76.58 & 87.48 & 69.86 & 71.11 & 81.66 & 73.72 & 56.30 & 86.07 & 78.68 & 59.22 & 78.59 & 73.22 \\ 
        %DANN~\cite{ganin2016domain} & 56.17 & 81.72 & 86.87 & 68.67 & 73.38 & 83.76 & 69.92 & 56.84 & 85.80 & 79.41 & 57.26 & 78.26 & 73.17 \\ 
        %RTN~\cite{long2016unsupervised} & 50.46 & 77.80 & 86.90 & 65.12 & 73.40 & 85.07 & 67.86 & 45.23 & 85.50 & 79.20 & 55.55 & 78.79 & 70.91 \\ 
        IWAN~\cite{zhang2018importance} & \xmark &  52.55 & 81.40 & 86.51 & 70.58 & 70.99 & 85.29 & 74.88 & 57.33 & 85.07 & 77.48 & 59.65 & 78.91 & 73.39 \\ 
        PADA~\cite{zhang2018importance} & \xmark & 39.58 & 69.37 & 76.26 & 62.57 & 67.39 & 77.47 & 48.39 & 35.79 & 79.60 & 75.94 & 44.50 & 78.10 & 62.91 \\ 
        ATI~\cite{panareda2017open}
        % ~\cite{panareda2017open} 
        & \xmark & 52.90 & 80.37 & 85.91 & 71.08 & 72.41 & 84.39 & 74.28 & 57.84 & 85.61 & 76.06 & 60.17 & 78.42 & 73.29 \\ 
        OSBP~\cite{saito2018open} & \xmark & 47.75 & 60.90 & 76.78 & 59.23 & 61.58 & 74.33 & 61.67 & 44.50 & 79.31 & 70.59 & 54.95 & 75.18 & 63.90 \\ 
        UAN~\cite{UDA_2019_CVPR} & \xmark & 63.00 & 82.83 & 87.85 & \textbf{76.88} & \textbf{78.70} & 85.36 & 78.22 & 58.59 & 86.80 & \textbf{83.37} & 63.17 & 79.43 & 77.02 \\
        % \midrule  \multicolumn{14}{c}{\textit{Source-free} adaptation}\\
        
        \hline
        
        Ours \textbf{\textit{USFDA}} & \cmark & \textbf{63.35} & \textbf{83.30} & \textbf{89.35} & 70.96 & 72.34 & \textbf{86.09} & \textbf{78.53} & \textbf{60.15} & \textbf{87.35} & 81.56 & 63.17 & \textbf{88.23} & \textbf{77.03} \\
        % Ours \textbf{\textit{USFDA-b}} & \cmark & 62.46 & 82.71 & 88.26 & 71.10 & 70.88 & 85.75 & 78.21 & 59.18 & 86.05 & 82.17 & \textbf{63.22} & 87.68 & 76.47 \\
        \bottomrule
        
    \end{tabular}\vspace{-6mm}
    }
\end{table*}

%\textbf{4.2\; Evaluation Methodology}

%%%%%%%%%% figure 3 %%%%%%%%%%%%%%%%%%
\begin{figure*}[!t]
    \centering
    \includegraphics[width=1\linewidth]{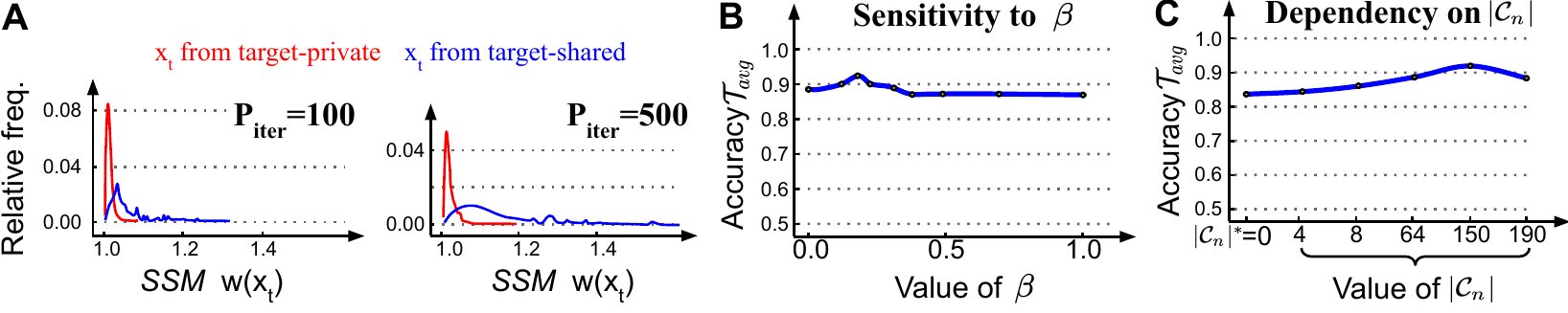}\vspace{-4mm}
    \caption{\small Ablative analysis on the task \textbf{A}$\rightarrow$\textbf{D} ({Office-31}). \textbf{A)} Histogram of SSM values of $x_t$ separately for target-private and target-shared samples at the \textit{Procurement} iteration 100 (top) and 500 (bottom). \textbf{B)} The sensitivity curve for $\beta$ shows marginally stable adaptation accuracy for a wide-range of values. \textbf{C)} A marginal increase in $\mathcal{T}_{avg}$ is observed with increase in $\vert\mathcal{C}_n\vert$.}
    \label{fig_ssm_sensitivity}\vspace{-1mm}
\end{figure*}
%%%%%%%%%%%%%%%%%%%%%%%%%%%%%%%%%%%%%%

% \vspace{-1mm}
\subsection{Evaluation Methodology}
% \vspace{-1mm}

\textbf{a) Average accuracy on Target dataset, $\mathcal{T}_{avg}$.} We resort to the evaluation protocol proposed in the VisDA2018 Open-Set Classification challenge. Accordingly, all the target-private classes are grouped into a single "\textit{unknown}" class and the metric reports the average of per-class accuracy over {$\vert\mathcal{C}_s\vert+1$} classes. In our framework, a target sample is marked as "\textit{unknown}" if it is classified (${\textit{argmax}}_k\hat{z}^{(k)}$) into any of the negative $\vert{\mathcal{C}_n}\vert$ classes. 
% out of total $\vert{\mathcal{C}_s}\vert+\vert{\mathcal{C}_n}\vert$ categories. 
In contrast, UAN~\cite{UDA_2019_CVPR} relies on the sample-level weight, to mark a target sample as "\textit{unknown}" based on a sensitive threshold hyperparameter.
% In contrast, UAN~\cite{UDA_2019_CVPR} relies on a sensitive hyperparameter, as a threshold on the sample-level weighting, to mark a target sample as "\textit{unknown}". 
Also note that our method is truly \textit{source-free} during adaptation, while all other methods have access to the full source-data.

\vspace{1mm}
\textbf{b) Accuracy on Target-Unknown data, $\mathcal{T}_{unk}$.} We evaluate the target unknown accuracy, $\mathcal{T}_{unk}$, as the proportion of actual target-private samples (i.e. $\{(x_t,y_t):y_t\in\overline{\mathcal{C}}_t\}$) being classified as "\textit{unknown}" after adaptation. Note that, UAN~\cite{UDA_2019_CVPR} does not report $\mathcal{T}_{unk}$ which is a crucial metric to evaluate the vulnerability of the model after its deployment in the target environment. The $\mathcal{T}_{avg}$ metric fails to capture this as a result of class-imbalance in the \textit{Open-set} scenario~\cite{saito2018open}. Hence, to realize a common evaluation ground, we train the UAN implementation %\footnote{Implementation of UAN%~\cite{UDA_2019_CVPR}
%: \url{https://github.com/thuml/Universal-Domain-Adaptation}.} 
provided by the authors~\cite{UDA_2019_CVPR} and denote it as UAN* in further sections of this paper. 
We observe that, the UAN\cite{UDA_2019_CVPR} training algorithm is often unstable with a decreasing trend of $\mathcal{T}_{unk}$ and $\mathcal{T}_{avg}$ over increasing training iterations. We thus report the mean and standard deviation of the peak values of $\mathcal{T}_{unk}$ and $\mathcal{T}_{avg}$ achieved by UAN*, over 5 separate runs on Office-31 dataset (see Table~\ref{tab:table_2}). 

\vspace{1mm}
\textbf{c) Implementation Details.} We implement our network in PyTorch
% \footnote{Implementation: \textbf{\url{https://github.com/val-iisc/usfda}}} 
and use ResNet-50~\cite{he2016deep} as the backbone-model $M$, pre-trained on ImageNet~\cite{imagenet} inline with UAN~\cite{UDA_2019_CVPR}. The complete architecture of other components is provided in the Supplementary. We denote our approach as \textit{USFDA}. A sensitivity analysis of the major hyper-parameters used in the proposed framework is provided in Fig.~\ref{fig_ssm_sensitivity}{\color{red}B-C}, and Suppl. Fig. {\color{red}2B}. In all our ablations across the datasets, we fix the hyperparameters values as $\alpha=0.2$ and $\beta=0.1$. We utilize Adam optimizer~\cite{kingma2014adam} with a fixed learning rate of $0.0001$ for training in both the \textit{Procurement} and the \textit{Deployment} stages. For the implementation of UAN*, we use the hyper-parameter value $w_{0}=-0.5$, as specified by the authors for the task \textbf{A}$\rightarrow$\textbf{D} in the Office-31 dataset. 

%\textbf{4.3\; Discussion}

% \vspace{-1mm}
\subsection{Discussion}
% \vspace{-1mm}
\label{sec:discussion}

\textbf{a) Comparison against prior arts.} We compare our approach with UAN~\cite{UDA_2019_CVPR}, and other prior methods. The results are presented in Tables~\ref{tab:table_1}{\color{red}-}\ref{tab:table_2}. Our approach yields state-of-the-art results even in a \textit{source-free} setting on several tasks. Particularly in Table~\ref{tab:table_2}, we present $\mathcal{T}_{unk}$ on various datasets and also report the mean and standard-deviation for both the accuracy metrics computed over 5 random initializations in the Office-31 dataset (the last six rows). Our method is able to achieve much higher $\mathcal{T}_{unk}$ than UAN*~\cite{UDA_2019_CVPR}, highlighting our superiority as a result of the novel learning approach incorporated in both \textit{Procurement} and \textit{Deployment} stages. 
% Note, both \textit{USFDA-a} and \textit{USFDA-b} yield similar results across a wide range of standard benchmarks. 
We also perform a characteristic comparison of algorithm complexity in terms of the amount of learnable parameters and training time; a) \textit{Procurement}: [11.1M, 380s], b) \textit{Deployment}: [3.5M, 44s], c) UAN~\cite{UDA_2019_CVPR}: [26.7M, 450s] ({in a consistent setting}). The significant computational advantage in the \textit{Deployment} stage makes our approach highly suitable for real-time adaptation.
In contrast to UAN, the proposed framework offers a much simpler adaptation algorithm
devoid of networks such as an adversarial discriminator and additional finetuning of the ResNet-50 backbone. %(+26M parameters in UAN~\cite{UDA_2019_CVPR}),
\begin{table*}[!t]
    \addtolength{\tabcolsep}{-2.4pt}
    \renewcommand{\arraystretch}{1.3}
    \centering
    \tiny
    \caption{\small$\mathcal{T}_{avg}$ on \textbf{Office-31} (with $\vert{\mathcal{C}}\vert / \vert{\mathcal{C}_s\cup\mathcal{C}_t\vert}=0.32$), \textbf{VisDA} (with $\vert{\mathcal{C}}\vert / \vert{\mathcal{C}_s\cup\mathcal{C}_t\vert}=0.50$), and \textbf{ImageNet-Caltech} (with $\vert{\mathcal{C}}\vert / \vert{\mathcal{C}_s\cup\mathcal{C}_t\vert}=0.07$). Scores for the prior works are directly taken from UAN~\cite{UDA_2019_CVPR}. SF denotes support for \textit{source-free} adaptation. \vspace{-6mm}}
    \label{tab:table_2}
    \resizebox{1\textwidth}{!}{
    \begin{tabular}{c|c|ccccccc|c|cc}
        \toprule
        \multirow{2}{20pt}{\centering Method}& \multirow{2}{*}{\centering SF} &  \multicolumn{7}{c|}{Office-31} & \multicolumn{1}{c|}{VisDA} & \multicolumn{2}{c}{ImNet-Caltech} \\
        \cmidrule(lr){3-9}\cmidrule(lr){10-10}\cmidrule(lr){11-12}
        && A$\rightarrow$W & D$\rightarrow$W & W$\rightarrow$D & A$\rightarrow$D & D$\rightarrow$A & W$\rightarrow$A & Avg & S $\rightarrow$ R & I $\rightarrow$ C & C $\rightarrow$ I \\
        \hline\hline
        ResNet~\cite{he2016deep}&\xmark & 75.94 & 89.60 & 90.91 & 80.45 & 78.83 & 81.42 & 82.86 & 52.80 & 70.28 & 65.14 \\
        %DANN~\cite{ganin2016domain}&\xmark & 80.65 & 80.94 & 88.07 & 82.67 & 74.82 & 83.54 & 81.78 & 52.94 & 71.37 & 66.54 \\
        %RTN~\cite{long2016unsupervised}&\xmark & 85.70 & 87.80 & 88.91 & 82.69 & 74.64 & 83.26 & 84.18 & 53.92 & 71.94 & 66.15\\
        IWAN~\cite{zhang2018importance}&\xmark & 85.25 & 90.09 & 90.00 & 84.27 & 84.22 & 86.25 & 86.68 & 58.72 & 72.19 & 66.48\\
        PADA~\cite{zhang2018importance}&\xmark & 85.37 & 79.26 & 90.91 & 81.68 & 55.32 & 82.61 & 79.19 & 44.98 & 65.47 & 58.73\\
        ATI ~\cite{panareda2017open}
        &\xmark & 79.38 & 92.60 & 90.08 & 84.40 & 78.85 & 81.57 & 84.48 & 54.81 & 71.59 & 67.36\\
        OSBP~\cite{saito2018open}&\xmark & 66.13 & 73.57 & 85.62 & 72.92 & 47.35 & 60.48 & 67.68 & 30.26 & 62.08 & 55.48\\
        UAN~\cite{UDA_2019_CVPR}&\xmark & 85.62 & 94.77 & 97.99 & 86.50 & 85.45 & 85.12 & 89.24 & 60.83 & 75.28 & 70.17\\
        
        \hline
        
        UAN* $\mathcal{T}_{avg}$ &\xmark& 83.00$\pm$1.8 & 94.17$\pm$0.3 & 95.40$\pm$0.5 & 83.43$\pm$0.7 & 86.90$\pm$1.0 & \textbf{87.18}$\pm$0.6 & 88.34 & 54.21%$\pm$4.72 
        & 74.77 & 71.51\\
        
        Ours \textbf{\textit{USFDA}} $\mathcal{T}_{avg}$ &\cmark& \textbf{85.56}$\pm$1.6 & 95.20$\pm$0.3 & \textbf{97.79}$\pm$0.1 & \textbf{88.47}$\pm$0.3 & 87.50$\pm$0.9 & 86.61$\pm$0.6 & \textbf{90.18} & \textbf{63.92}%$\pm$2.31 \\
        & \textbf{76.85} & 72.13\\
        
        % Ours \textbf{\textit{USFDA-b}} $\mathcal{T}_{avg}$ &\cmark& 83.21$\pm$1.2 & \textbf{95.33}$\pm$0.3 & 96.37$\pm$0.3 & 86.84$\pm$0.4 & \textbf{87.91}$\pm$0.6 & 86.74$\pm$0.5 & 89.40 
        % & 62.77%$\pm$2.31 \\
        % & 76.74 & \textbf{72.25}\\
        
        \hline
        
        UAN* $\mathcal{T}_{unk}$ &\xmark& 20.72$\pm$11.7 & 53.53$\pm$2.4 & 51.57$\pm$5.0 & 34.43$\pm$3.3 & 51.88$\pm$4.8 & 43.11$\pm$1.3 & 42.54 & 19.68%$\pm$9.56 \\
        & 33.43 & 31.24\\
        
        Ours \textbf{\textit{USFDA}} $\mathcal{T}_{unk}$ &\cmark& \textbf{73.98}$\pm$7.5 & 85.64$\pm$2.2 & \textbf{80.00}$\pm$1.1 & 82.23$\pm$2.7 & \textbf{78.59}$\pm$3.2 & \textbf{75.52}$\pm$1.5 & \textbf{79.32} & \textbf{36.25}%$\pm$3.89 \\
         & \textbf{51.21} & \textbf{48.76}\\
         
        %  Ours \textbf{\textit{USFDA-b}} $\mathcal{T}_{unk}$ &\cmark& 70.22$\pm$8.8 & \textbf{85.89}$\pm$2.3 & 78.29$\pm$1.7 & \textbf{84.66}$\pm$3.1 & 76.22$\pm$2.8 & 73.91$\pm$1.6 & 78.19 & 34.84%$\pm$3.89 \\
        %  & 51.10 & 48.20\\
        
        \bottomrule
    \end{tabular}%
    }
    % \vspace{-2mm}
\end{table*}
%%%%%%%%%%%%%%%%%%%%%%% end %%%%%%%%%%%%%%%%%%%%%%%%%%%%

%%%%%%%%%% figure 2 %%%%%%%%%%%%%%%%%%
\begin{figure*}[!t]
    \centering
    \includegraphics[width=1\linewidth]{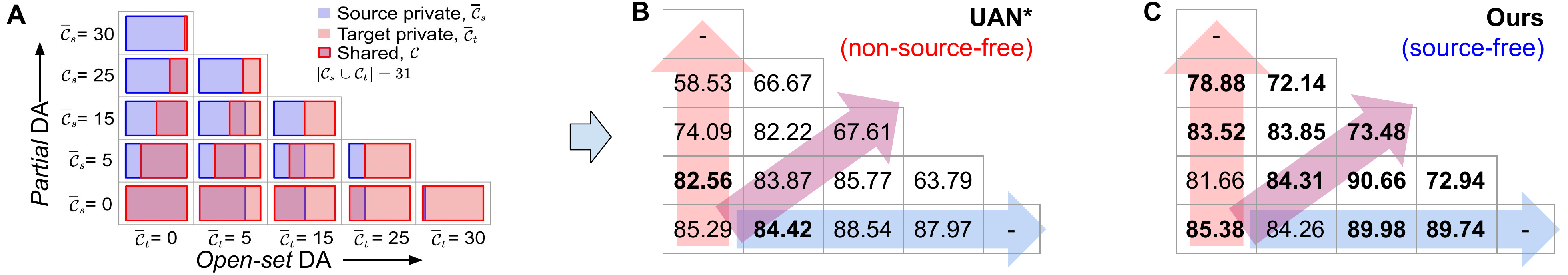}
    \vspace{-6mm}
    \caption{\small Comparison across varied label-set relationships for the task \textbf{A}$\rightarrow$\textbf{D} in {Office-31} dataset. \textbf{A)} Visual representation of label-set relationships and $\mathcal{T}_{avg}$ at the corresponding instances for \textbf{B)} UAN*~\cite{UDA_2019_CVPR} and \textbf{C)} ours \textit{source-free} model. Effectively, the direction along x-axis (blue horizontal arrow) characterizes increasing \textit{Open-set} complexity. The direction along y-axis (red vertical arrow) shows increasing complexity of \textit{Partial} DA scenario. The pink diagonal arrow denotes the effect of decreasing shared label space.}
    \label{fig_category_gap_table}
    \vspace{-2mm}
\end{figure*}
%%%%%%%%%%%%%%%%%%%%%%%%%%%%%%%%%%%%%%

% %%%%%%%%%% figure 3 %%%%%%%%%%%%%%%%%%
% \begin{figure*}[!t]
%     \centering
%     \includegraphics[width=1.0\columnwidth]{figures/iclr_uda_fig_3b.pdf}%\vspace{-1mm}
%     \caption{\small\textbf{A.} Sensitivity against $\vert\mathcal{C}_n\vert$, represented by $\vert\mathcal{C}_n\vert/^{\vert\mathcal{C}_s\vert}C_2$ for varying $\vert\overline{\mathcal{C}}_s\vert$ or $\vert\overline{\mathcal{C}}_t\vert$ (see fig. legend) by fixing the others (top cyan box), across varied datasets. \textbf{B.} Sensitivity against $\alpha$ and batch-size ratio (fixed $b_{+ve}+b_{-ve}=64$). Note the scale of Y-axis.%of $\mathcal{T}_{avg}$.
%     % across varied datasets.
%     }
%     \label{fig_5}\vspace{-1mm}
% \end{figure*}
% %%%%%%%%%%%%%%%%%%%%%%%%%%%%%%%%%%%%%%

\vspace{0.5mm}
\textbf{b) Does \textit{SSM} satisfy the expected inequality?} Effectiveness of the proposed learning algorithm, in case of \textit{source-free} deployment, relies on the formulation of \textit{SSM}, which is expected to satisfy Eq.~\ref{eqn_1}. Fig.~\ref{fig_ssm_sensitivity}{\color{red}A} shows a histogram of the \textit{SSM} separately for samples from target-shared (blue) and target-private (red) label space. The success of this metric is attributed to the generative nature of \textit{Procurement} stage, which enables the source model to distinguish between the marginally more negative target-private samples as compared to the samples from the shared label space.

%Fig. 4B shows the the probability density function of the weight $w(x_t)$ calculated for target-shared (red) and target-unknown (orange) samples. Due to the generative nature of procurement stage, the source classifier is more robust to out-of-distribution samples and produces lesser confidence predictions for the samples from target-unknown classes, thereby creating a division in $w(x_t)$.

\vspace{0.5mm}
\textbf{c) Sensitivity to hyper-parameters.} As we tackle DA in a \textit{source-free} setting simultaneously intending to generalize across varied \textit{category-gaps}, a low sensitivity to hyperparameters would further enhance our practical usability. To this end, we fix certain hyperparameters for all our experiments (also in Fig.~\ref{fig_category_gap_table}{\color{red}C}) even across datasets (i.e. $\alpha=0.2$, $\beta=0.1$). Thus, one can treat them as global-constants with $\vert\mathcal{C}_n\vert$ being the only hyperparameter, as variations in one by fixing the others yield complementary effect on regularization in the \textit{Procurement} stage.  %(i.e. except $\beta$). 
A thorough analysis reported in the Suppl. Fig. {\color{red}2}, demonstrates a reasonably low sensitivity of our model to these hyperparameters. 

%A sensitivity analysis of the major hyper-parameters used in the proposed framework is provided in Fig.~\ref{fig_4}{\color{red}B}.

\vspace{0.5mm}
\textbf{d) Generalization across category-gap.} One of the key objectives of the proposed framework is to effectively operate in the absence of the knowledge of label-set relationships. To evaluate it in the most compelling manner, we propose a tabular form shown in Fig.~\ref{fig_category_gap_table}{\color{red}A}. We vary the number of private classes for target and source along the x-axis and y-axis respectively, with a fixed $\vert\mathcal{C}_s\cup\mathcal{C}_t\vert=31$. We compare the $\mathcal{T}_{avg}$ metric at the corresponding table instances, shown in Fig.~\ref{fig_category_gap_table}{\color{red}B-C}. The results clearly highlight superiority of the proposed framework specifically for the more practical scenarios (close to the diagonal instances) as compared to the unrealistic \textit{Closed-set} setting ($\vert{\overline{\mathcal{C}}_s}\vert=\vert{\overline{\mathcal{C}}_t}\vert=0$).

\vspace{0.5mm}
\textbf{e) DA in absence of shared categories.}
In universal adaptation, we seek to transfer the knowledge of "\textit{class-separability criterion}" obtained from the source domain to the deployed target environment. More concretely, it is attributed to the segregation of data samples based on some expected characteristics, such as classification of objects according to their pose, color, or shape etc. To quantify this, we consider an extreme case where $\mathcal{C}_s\cap\mathcal{C}_t=\emptyset$  (\textbf{A}$\rightarrow$\textbf{D} in Office-31 with $\vert\mathcal{C}_s\vert=15$, $\vert\mathcal{C}_t\vert=16$). Allowing access to a single labeled target sample from each category in $\overline{\mathcal{C}}_t=\mathcal{C}_t$, we aim to obtain a one-shot recognition accuracy (assignment of cluster index or class label using the one-shot samples as the cluster center at $F_t\circ M(x_t)$)
%(Nearest-neighbour classification at $u$) 
to quantify the above metric.
%predicted label $=\text{argmin}_k(||a_{t} - a_{c_k}||_2)$) to quantify the above metric. 
We obtain 64.72\% accuracy for the proposed framework as compared to 13.43\% for UAN*.
%A significant improvement in accuracy over UAN*~\cite{UDA_2019_CVPR} 
This strongly validates our superior knowledge transfer capability as a result of the generative classifier with labeled negative samples complementing for the target-private categories.

%indicates that, the samples are inherently clustered in the intermediate feature level validating an efficient transfer of “\textit{class separability}” in a fully unsupervised manner.
%, demonstrating superior knowledge transfer capability.

% \textbf{Varying the number of shared and unknown classes.} We also compare $\mathcal{T}_{avg}$ on varying the number of unknown classes $\vert{\overline{\mathcal{C}}_s}\vert$ and $\vert{\overline{\mathcal{C}}_t}\vert$, on Office-31 dataset, following $\vert{\overline{\mathcal{C}}_s}\vert + \vert{\overline{\mathcal{C}}_t}\vert + \vert{\mathcal{C}}\vert = 31$. Comparing UAN and our model, TODO: conclude the results. \textbf{TODO: Regarding figure 3:} accuracy of our model with |C| = 1 (diagonal elements), is higher than UAN because of the negative image conditioning

\vspace{0.5mm}
\textbf{f) Dependency on the simulated negative dataset.} Conceding that a combinatorial amount of negative labels can be created, we evaluate the scalability of the proposed approach, by varying the number of negative classes in the \textit{Procurement} stage by selecting $0$, $4$, $8$, $64$, $150$ and $190$ negative classes as reported in the X-axis of Fig.~\ref{fig_ssm_sensitivity}{\color{red}C}. For the case of $0$ negative classes, denoted as $\vert\mathcal{C}_n\vert^{*}=0$ in Fig.~\ref{fig_ssm_sensitivity}{\color{red}C}, we synthetically generate random negative features at the intermediate level $u$, which are at least {3}-$\sigma$ away from each of the positive source priors $P(u_s|c_i)$ for $i = 1, 2, ..., |\mathcal{C}_s|$. We then make use of these feature samples along with positive image samples, to train a $(\vert\mathcal{C}_s\vert+1)$ class \textit{Procurement} model with a single negative class. The results are reported in Fig.~\ref{fig_ssm_sensitivity}{\color{red}C} on the {\textbf{A}$\rightarrow$\textbf{D}} task of Office-31 dataset with category relationship inline with the setting in Table~\ref{tab:table_2}. We observe an acceptable drop in accuracy with decrease in number of negative classes, hence validating scalability of the approach for large-scale classification datasets (such as ImageNet). 
% Similarly, we also tried combining three or more images to form such negative classes. 
Similarly, we also evaluated our framework by combining three or more images to form such negative classes.
However, we found that with increasing number of negative classes ($\permcomb{C}{\vert\mathcal{C}_s\vert}{3} > \permcomb{C}{\vert\mathcal{C}_s\vert}{2}$), the model achieves under-fitting on positive source categories (similar to Fig.~\ref{fig_ssm_sensitivity}{\color{red}C}, where accuracy reduces beyond a certain limit because of over regularization). 

\section{Conclusion}
%\vspace{-2mm}
We have introduced a novel Universal \textit{Source-Free} Domain Adaptation framework, acknowledging practical domain adaptation scenarios devoid of any assumption on the source-target label-set relationship. In the proposed two-stage framework, learning in the \textit{Procurement} stage is found to be highly crucial, as it aims to exploit the knowledge of class-separability in the most general form with enhanced robustness to out-of-distribution samples. Besides this, the success in the \textit{Deployment} stage is attributed to the well-designed learning objectives effectively utilizing the source similarity criterion. This work can be served as a pilot study towards learning efficient inheritable models in future.

\vspace{1mm}\noindent
\textbf{Acknowledgements.} This work is supported by a Wipro PhD Fellowship (Jogendra) and a grant from Uchhatar Avishkar Yojana (UAY, IISC\_010), MHRD, Govt. of India. 
% We also thank Ujjawal Sharma (IIT Roorkee) for helping out with the experiments.
We would also like to thank Ujjawal Sharma (IIT Roorkee) for assisting with the implementation of prior arts.

% \newpage
{\small
\bibliographystyle{ieee_fullname}
\bibliography{references}
}



\section*{Supplementary material}

# Supplementary: Universal Source-Free Domain Adaptation

This Supplementary is organized as follows,

- Sec. 1: Notations
- Sec. 2: Implementation details
  - *Procurement* Stage. (Sec. 2.1, Algo. 1)
  - *Deployment* Stage. (Sec. 2.2)
- Sec. 3: Additional Results
  - Pretraining the backbone network on Places instead of ImageNet. (Sec. 3.1, Table 2)
  - Space and Time complexity analysis. (Sec. 3.2)
  - Varying label-set relationship. (Sec. 3.3, Fig. 1)
  - Sensitivity analysis. (Sec. 3.4, Fig. 2)
  - Closed-set adaptation. (Sec. 3.5, Table 3)
  - Accuracy on source dataset after *Procurement*. (Sec. 3.6)
  - Incremental one-shot classification. (Sec. 3.7)
  - Feature space visualization. (Sec. 3.8, Fig. 3)
- Sec. 4: Miscellaneous
  - Specifications of computing resources. (Sec. 4.1)
  - References to code. (Sec. 4.2)

## 1. Notations

We summarize the notations used in the paper in Table. 1.

## 2. Implementation Details

Here, we describe the network architectures and the training process used for the *Procurement* and the *Deployment* stages of our approach.

### 2.1. Procurement Stage

**a) Design of the classifier $D$.** Keeping in mind the possibility of the model encountering an additional domain shift after having adapted from the source domain to a target domain (e.g. encountering domain **W** after performing the adaptation **A** → **D** in **Office-31** dataset), we design the classifier's architecture in a manner which allows for modification in the number of negative classes post *Procurement*.

Table 1: Notation Table

| | Symbol | Description |
|---|---|---|
| Distribution | $p$ | Marginal source input distribution |
| | $p_n$ | Marginal negative feature distribution |
| | $q$ | Marginal target input distribution |
| | $p_{\bar{s}}$ | Marginal source-private distribution |
| | $q_{\bar{t}}$ | Marginal target-private distribution |
| | $P(u_s\|c_i)$ | Gaussian prior for the source samples |
| Network | $M$ | Backbone model |
| | $F_s$ | Source feature extractor |
| | $F_t$ | Target feature extractor |
| | $G$ | Decoder |
| | $D$ | Classifier |
| Sets | $\mathcal{D}_s$ | Labeled source dataset |
| | $\mathcal{D}_n$ | Labeled negative dataset |
| | $\mathcal{D}_t$ | Unlabelled target dataset |
| | $\mathcal{C}_s$ | Label-set of the source domain |
| | $\mathcal{C}_n$ | Label-set of the negative samples |
| | $\mathcal{C}_t$ | Label-set of the target domain |
| | $\mathcal{C}$ | Shared label-set |
| | $\overline{\mathcal{C}}_s$ | Source-private label-set |
| | $\overline{\mathcal{C}}_t$ | Target-private label-set |
| Samples / Misc. | $(x_s, y_s)$ | Paired source samples |
| | $(x_n, y_n)$ | Paired negative samples |
| | $x_t$ | Unpaired target samples |
| | $v_s, v_t$ | Output of $M$ for source / target resp. |
| | $\hat{v}_s$ | Output of $G$ |
| | $u_s, u_t$ | Output of $F_s$ / $F_t$ resp. |
| | $u_r$ | Samples drawn from class priors |
| | $\mu_{c_i}$ | Mean feature $u_s$ for $c_i \in \mathcal{C}_s$ |
| | $\Sigma_{c_i}$ | Covariance of $u_s$ for $c_i \in \mathcal{C}_s$ |
| | $d$ | Output of $D$ (logit vector) |
| | $\sigma^{(k)}(\cdot)$ | $k^{\text{th}}$ element of the softmax vector |
| | $\hat{z}$ | Softmax over $\|\mathcal{C}_s\| + \|\mathcal{C}_n\|$ logits |
| | $\tilde{z}_s$ | Softmax over $\|\mathcal{C}_s\|$ logits |
| | $\tilde{z}_n$ | Softmax over $\|\mathcal{C}_n\|$ logits |
| | $w(\cdot), w'(\cdot)$ | SSM and its complement resp. |

**Algorithm 1** Negative dataset generation using Image-composition

1: **input:** Image pair $(I_1, I_2) \in \mathcal{D}_s$. (image shape $H$x$W$x$3$ = 224x224x3)
2: $k \leftarrow 30$
3: $x_1, x_2, y_1, y_2 \leftarrow \text{rand}(0, W), \text{rand}(0, W), \text{rand}(0, H), \text{rand}(0, H)$
4: $c_x, c_y \leftarrow \text{rand}(W/2 - k, W/2 + k), \text{rand}(H/2 - k/3, H/2 + k/3)$
5: $d_x, d_y \leftarrow \text{rand}(W/2 - k/3, W/2 + k/3), \text{rand}(H/2 - k, H/2 + k)$

  **Horizontal Splicing**
6: $s_1 \leftarrow \text{quadratic\_interpolation}([(0, y_1), (c_x, c_y), (223, y_2)])$

  **Vertical Splicing**
7: $s_2 \leftarrow \text{quadratic\_interpolation}([(x_1, 0), (d_x, d_y), (x_2, 223)])$

  **Defining Masks**
8: $m_1 \leftarrow$ mask region below $s_1$
9: $m_2 \leftarrow$ mask region to the left of $s_2$

  **Merging alternate regions to create composite images**
10: $I_a \leftarrow m_1 * I_1 + (1 - m_1) * I_2$
11: $I_b \leftarrow m_2 * I_1 + (1 - m_2) * I_2$
12: $I_c \leftarrow m_1 * I_2 + (1 - m_1) * I_1$
13: $I_d \leftarrow m_2 * I_2 + (1 - m_2) * I_1$

14: **return** $I_a, I_b, I_c, I_d$

We achieve this by maintaining two separate classifiers during the *Procurement* stage - $D_{src}$ that operates on the positive source classes, and, $D_{neg}$ that operates on the negative source classes (see architecture in Table 5). The final classification score is obtained by computing softmax over the concatenation of logit vectors produced by $D_{src}$ and $D_{neg}$. Therefore, the model can be retrained on a different number of negative classes post *Deployment* (using another negative class classifier $D_{neg}^2$), thus preparing it for a subsequent adaptation step to another domain.

**b) Negative dataset generation.** We generate the negative dataset $\mathcal{D}_n$ by compositing images taken from different positive source classes, as described in Algo. 1. We generate random masks using quadratic splines passing through a central image region (lines 3-9). Using these masks, we merge alternate regions of the images, both horizontally and vertically, resulting in 4 negative images for each pair of images (lines 10-13). To effectively cover the inter-class negative region, we randomly sample image pairs from $D_s$ belonging to different classes, however we do not impose any constraint on how the classes are selected (for e.g. one can composite images from an animal and a non-animal class). We choose 5000 pairs for tasks on **Office-31**, **Office-Home** and **VisDA** datasets, and 12000 for **ImageNet-Caltech**. Since the input source distribution ($p$) is fixed we first synthesize a negative dataset offline (instead of creating them on the fly) to ensure finiteness of the training set. The training algorithm for *USFDA* is given in Algo. 1 of the paper.

**c) Justification of $\mathcal{L}_p$.** The cross-entropy loss enforced on the likelihoods (referred to as $\mathcal{L}_p$ in the paper) not only enforces intra-class compactness but also ensures inter-class separability in the latent $u$-space. Since the negative samples are only an approximation of the future target private classes that are expected to be encountered, we choose not to employ this loss for negative samples. Such a training procedure, eventually results in a natural development of bias towards the confident positive source classes. This subsequently leads to the placement of source clusters in a manner which enables source-free adaptation (See Fig. 4C of the paper).

**d) Minibatch negative sampling strategy.** We create an unbiased batch of training samples for a training iteration by sampling equal number of positive and negative samples from the dataset. Particularly, we sample 32 positive source class images ($b_{+ve} = 32$) and 32 negative images ($b_{-ve} = 32$) for each training iteration. This gives an effective batch size of $b_{+ve} + b_{-ve} = 64$.

**e) Use of multiple optimizers for training.** In the presence of multiple losses, we subvert a time-consuming loss-weighting hyperparameter search by making use of multiple Adam optimizers during training. Essentially, we define a separate optimizer for each loss term, and optimize only one of the losses (chosen in a round robin fashion) in each iteration of training. We use a learning rate of 0.0001 for each Adam optimizer. Intuitively, the moment parameters in each Adam optimizer adaptively scales the corresponding gradients, thereby avoiding loss-scaling hyperparameters.

**f) Label-Set Relationships.** For **Office-31** dataset in the UDA setting, we use the 10 classes shared by Office-31 [5] and Caltech-256 [1] as the shared label-set $\mathcal{C}$. These classes are: *back_pack, calculator, keyboard, monitor, mouse, mug, bike, laptop_computer, headphones, projector*. From the remaining classes, in alphabetical order, we choose the first 10 classes as source-private ($\overline{\mathcal{C}}_s$) classes, and the rest 11 as target-private ($\overline{\mathcal{C}}_t$) classes. For **VisDA**, alphabetically, the first 6 classes are considered as $\mathcal{C}$, the next 3 as $\overline{\mathcal{C}}_s$ and the last 3 comprise $\overline{\mathcal{C}}_t$. The **Office-Home** dataset has 65 categories, of which we use the first 10 classes as $\mathcal{C}$, the next 5 for $\overline{\mathcal{C}}_s$, and the rest 50 classes as $\overline{\mathcal{C}}_t$.

## 2.2. Deployment Stage

**a) Architecture.** The network architecture used during the *Deployment* stage is given in Table 6. Note that the decoder $G$ used during the *Procurement* stage, is not available during *Deployment*, restricting complete access to the source data.

**b) Training.** The only trainable component is the Feature Extractor $F_t$, which is initialized from $F_s$. Here, the *SSM* is calculated by passing the target images through the network trained on source data (source model), i.e for each image $x_t$, we calculate $\hat{y} = \sigma(D \circ F_s \circ M(x_t))$. Note that the softmax is calculated over all $|\mathcal{C}_s| + |\mathcal{C}_n|$ classes. This is done by concatenating the outputs of $D_{src}$ and $D_{neg}$, and then calculating softmax. Then, the *SSM* is determined by the exponential confidence of a target sample, where confidence is the highest softmax value in the categories in $\mathcal{C}_s$.

## 3. Additional Results

### 3.1. Pretraining the backbone network on Places instead of ImageNet.

We find that widely adopted standard domain adaptation datasets such as **Office-31** [5] and **VisDA** [4] often share a part or all of their label-set with **ImageNet**. Therefore, to validate our method's applicability when initialized from a network pretrained on an unrelated dataset, we attempt to solve the adaptation task **A→D** in **Office-31** dataset by pretraining the ResNet-50 backbone on **Places** dataset [8]. In Table 2 it can be observed that our method outperforms even source-dependent methods (e.g. UAN [7], which is also initialized a ResNet-50 backbone pretrained on **Places**). In contrast to our method, the algorithm in UAN [7] involves ResNet-50 finetuning. Therefore, we also compare against a variant of UAN with a frozen backbone network, by inserting an additional feature extractor that operates on the features extracted from ResNet-50 (similar to $F_s$ in the proposed method). The architecture of the feature extractor used for this variant of UAN is outlined in Table 4. We observe that our method significantly outperforms this variant of UAN with lesser number of trainable parameters (see Table 2).

Table 2: Evaluation of the proposed method on **A→D** task of **Office-31** [5] dataset, pretraining the ResNet-50 backbone ($M$) on Places instead of Imagenet. Note that, we set $|\mathcal{C}|/|\mathcal{C}_s \cup \mathcal{C}_t| = 0.32$, similiar to the setting used in Table 2 of the paper. Additionally, the last two columns of the table show a comparison between our method and UAN [7] with regard to the number of trainable parameters and total training time for adaptation.

| Method | ResNet-50 finetuning | Avg. per-class accuracy, $\mathcal{T}_{avg}$ | Number of Trainable Params. | Training time for Adaptation |
|---|---|---|---|---|
| UAN* | ✓ | 60.98 | 26.7 Million | 280s |
| UAN* | ✗ | 52.48 | 5.6 Million | 125s |
| **USFDA** | ✗ | **62.74** | **3.5 Million** | **44s** |

### 3.2. Space and Time complexity analysis.

On account of keeping the weights of the backbone network ($M$) frozen throughout the training process, and devoid of networks such as adversarial discriminator our method makes use of significantly lesser trainable parameters when compared to previous methods such as UAN [7] (See Table 2). Bereft of adversarial training, the proposed method also has a significantly lesser total training time for adaptation: 44 sec versus 280 sec in UAN (for the A→D task of Office-31 dataset and batch size of 32). Thus, the proposed framework offers a much simpler adaptation pipeline, with a superior computational complexity while achieving state-of-the-art domain adaptation performance across different datasets, even without accessing labeled source data at the time of adaptation (See Table 2). This corroborates the superiority of our method in real-time deployment scenarios.

### 3.3. Varying label-set relationship

In addition to the $\mathcal{T}_{avg}$ reported in Fig. 6 in the paper, we also compare the target-unknown accuracy $\mathcal{T}_{unk}$ for UAN* and USFDA. The results are presented in Fig. 1. Clearly, our method achieves a significant improvement over UAN on most settings. This demonstrates the capability of USFDA to detect outlier classes more efficiently, which can be attributed to the ingeniously developed *Procurement* stage.

### 3.4. Sensitivity Analysis

In all our experiments (across datasets as in Tables 1-2 of the paper and across varied label-set relationships as in Fig. 6 of the paper), we fix the hyperparameters as, $\alpha = 0.2$, $\beta = 0.1$, $|\mathcal{C}_n| = {}^{|\mathcal{C}_s|}C_2$ and $b_{+ve}/b_{-ve} = 1$. As mentioned in Sec. 4.3 of the paper, one can treat these hyperparameters as global constants. Nevertheless, in Fig. 2 we demonstrate the sensitivity of the model to these hyperparameters. Specifically, in Fig. 2A we show the sensitivity of the adaptation performance, to the choice of $|\mathcal{C}_n|$ during the *Procurement* stage, across a spectrum of label-set relationships. In Fig. 2B we show the sensitivity of the model to $\alpha$ and the batch-size

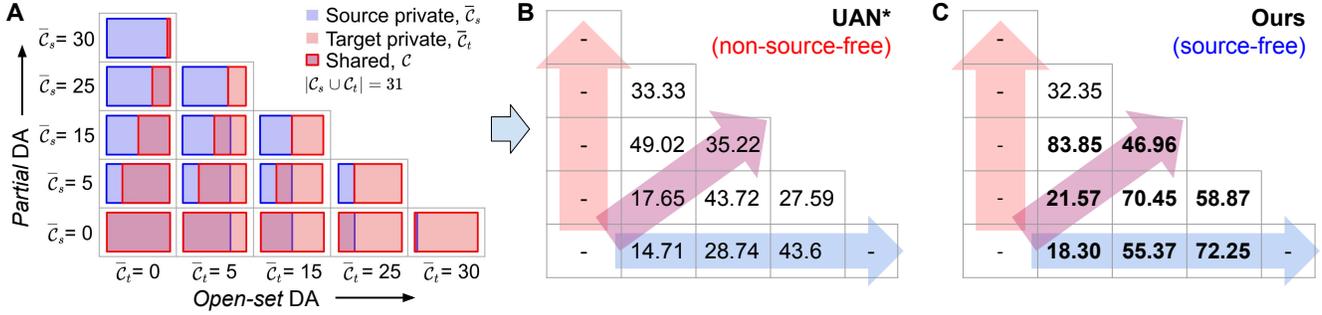

Figure 1: Comparison of $\mathcal{T}_{unk}$ across varied label-set relationships for the task A→D in Office-31 dataset. **A)** Visual representation of label-set relationships and $\mathcal{T}_{unk}$ at the corresponding instances for **B)** UAN* [7] and **C)** ours *source-free* model. Effectively, the direction along x-axis (blue horizontal arrow) characterizes increasing *Open-set* complexity. The direction along y-axis (red vertical arrow) shows increasing complexity of *Partial* DA scenario. And the pink diagonal arrow denotes the effect of decreasing shared label space.

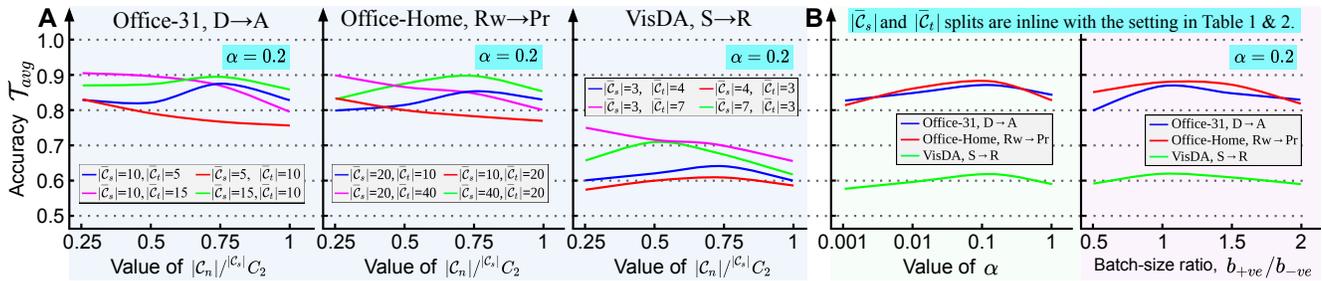

Figure 2: **A.** Sensitivity against $|\mathcal{C}_n|$, represented by $|\mathcal{C}_n|/|\mathcal{C}_s|C_2$ for varying $|\overline{\mathcal{C}}_s|$ or $|\overline{\mathcal{C}}_t|$ (see fig. legend) by fixing the others (top cyan box), across varied datasets. **B.** Sensitivity against $\alpha$ and batch-size ratio (fixed $b_{+ve} + b_{-ve} = 64$). Note the scale of X and Y-axis.

ratio $b_{+ve}/b_{-ve}$ (ratio of positive vs. negative samples during *Procurement*). Sensitivity to $\beta$ is shown in Fig. 5B of the paper. The model exhibits a reasonably low sensitivity to the hyperparameters, even in the challenging source-free scenario that allows for a reliable adaptation pipeline.

### 3.5. Closed-set adaptation

We additionally evaluate our method in the unsupervised source-free closed set adaptation scenario. In Table 3 we compare with the closed-set DA methods DAN [2], ADDA [6], CDAN [3] and the universal domain adaptation method UAN [7]. Note that, DAN, ADDA and CDAN rely on the assumption of a shared label space between the source and the target, and hence are not suited for a universal setting. Furthermore, all other methods require an explicit retraining on the source data during adaptation to perform well, even in the closed-set scenario. This clearly establishes the superiority of our method in the source-free setting.

### 3.6. Accuracy on source dataset after Procurement

We observe in our experiments that the accuracy on the source samples does not drop as a result of the partially generative framework. For the experiments conducted in Fig. 5C of the paper, we observe similar classification accuracy on the source validation set, on increasing the number of negative classes from 0 to 190. This effect can be attributed to a carefully chosen $\alpha = 0.2$, which is deliberately biased towards positive source samples to help maintain the discriminative power of the model even in the presence of class imbalance (i.e. $|\mathcal{C}_n| \gg |\mathcal{C}_s|$). This enhances the model's generative ability without compromising on the discriminative capacity on the positive source samples.

### 3.7. Incremental one-shot classification

In universal adaptation, we seek to transfer the knowledge of "*class separability*" obtained from the source domain to the deployed target environment. More concretely, it is attributed to the segregation of data samples based on an expected characteristics, such as classification of objects according to their pose, color, or shape etc. To quantify this, we consider an extreme case where $\mathcal{C}_s \cap \mathcal{C}_t = \emptyset$ (**A→D** in **Office-31** with $|\mathcal{C}_s| = 15$, $|\mathcal{C}_t| = 16$). Considering access to a single labeled target sample from each target category in $\overline{\mathcal{C}}_t = \mathcal{C}_t$, which are denoted as $x_t^{c_j}$, where $j = 1, 2, .., |\mathcal{C}_t|$, we perform one-shot Nearest-Neighbour based classification by obtaining the predicted class label as $\hat{c}_t = \text{argmin}_{c_j} ||F_t \circ M(x_t) - F_t \circ M(x_t^{c_j})||_2$. Then, the classification accuracy for the entire target set is computed by comparing $\hat{c}_t$ with

Table 3: Accuracy(%) on unsupervised closed-set DA (all use *ResNet50*). Ours is w/o hyperparmeter tuning. Refer Sec. 3.5.

| **Closed-set DA** methods | source-free | Universal-DA | **Office-31** | | | | | | | **VisDA** |
|---|---|---|---|---|---|---|---|---|---|---|
| | | | D→A | A→D | A→W | W→D | W→A | D→W | Avg. | S → R |
| DAN (ICML'15) | ✗ | ✗ | 63.6 | 78.6 | 80.5 | 99.6 | 62.8 | 97.1 | 80.4 | 61.1 |
| ADDA (CVPR'17) | ✗ | ✗ | 69.5 | 77.8 | 86.2 | 98.4 | 68.9 | 96.2 | 82.8 | - |
| CDAN (NeurIPS'18) | ✗ | ✗ | 70.1 | 89.8 | 93.1 | 100 | 68.0 | 98.2 | 86.5 | 66.8 |
| UAN (CVPR'19) | ✗ | ✓ | 68.4 | 85.3 | 81.2 | 99.1 | 69.7 | 98.1 | 83.6 | - |
| Ours *USFDA* | ✓ | ✓ | 70.4 | 85.4 | 81.6 | 98.0 | 69.4 | 98.4 | 83.9 | 59.8 |

the corresponding ground-truth category. We obtain 64.72% accuracy for the proposed framework as compared to 13.43% for UAN* [7]. A higher accuracy indicates that, the samples are inherently clustered in the intermediate feature level $M \circ F_t(x_t)$ validating an efficient transfer of "*class separability*" in a fully unsupervised manner.

### 3.8. Feature space visualization

We obtain a t-SNE plot at the intermediate feature level $u$ for both target and source samples (see Fig. 3), where the embedding for the target samples is obtained as $u_t = F_t \circ M(x_t)$ and the same for the source samples is obtained as $u_s = F_s \circ M(x_s)$. This is because we aim to learn domain-specific features in contrast to domain-agnostic features as a result of the restriction imposed by the source-free scenario ("cannot disturb placement of source clusters"). Firstly we obtain compact clusters for the source-categories as a result of the partially generative *Procurement* stage. Secondly, the target-private clusters are placed away from the source-shared and source-private as expected as a result of the carefully formalized *SSM* weighting scheme in the *Deployment* stage. This plot clearly validates our hypothesis.

## 4. Miscellaneous

### 4.1. Specifications of computing resources

For both *Procurement* and *Deployment* stages, we make use of the machine with the specifications as follows. CPU: Intel core i7-7700K, RAM: 32 GB, GPU: NVIDIA GeForce GTX 1080Ti (11 GB). The model is trained in Python 2.7 with PyTorch 1.0.0, with CUDA v8.0.61.

### 4.2. References to code

Our complete documented code (including data loaders, training pipeline etc.) used for the experiments is available at https://github.com/val-iisc/usfda. For evaluating UAN [7], we execute the official implementation provided by the authors on github[1].

Table 4: Architecture of the feature extractor used for UAN [7] under the "no ResNet-50 finetuning" case (see Table 2 and Sec. 3.1)

| **Operation** | **Features** | **Non-Linearity** |
|---|---|---|
| Input | 2048 | |
| Fully connected | 512 | ReLU |
| Fully connected | 256 | ReLU |
| Fully connected | 512 | ReLU |
| Fully connected | 2048 | ReLU |

## References

[1] Boqing Gong, Yuan Shi, Fei Sha, and Kristen Grauman. Geodesic flow kernel for unsupervised domain adaptation. In *CVPR*, 2012.

[2] Mingsheng Long, Yue Cao, Jianmin Wang, and Michael Jordan. Learning transferable features with deep adaptation networks. In *ICML*, 2015.

[3] Mingsheng Long, Zhangjie Cao, Jianmin Wang, and Michael I Jordan. Conditional adversarial domain adaptation. In *NeurIPS*, 2018.

[4] Xingchao Peng, Ben Usman, Neela Kaushik, Judy Hoffman, Dequan Wang, and Kate Saenko. Visda: The visual domain adaptation challenge. In *CVPR workshops*, 2018.

[5] Kate Saenko, Brian Kulis, Mario Fritz, and Trevor Darrell. Adapting visual category models to new domains. In *ECCV*, 2010.

[6] Eric Tzeng, Judy Hoffman, Kate Saenko, and Trevor Darrell. Adversarial discriminative domain adaptation. In *CVPR*, 2017.

[7] Kaichao You, Mingsheng Long, Zhangjie Cao, Jianmin Wang, and Michael I. Jordan. Universal domain adaptation. In *CVPR*, June 2019.

[8] Bolei Zhou, Agata Lapedriza, Aditya Khosla, Aude Oliva, and Antonio Torralba. Places: A 10 million image database for scene recognition. *IEEE Transactions on Pattern Analysis and Machine Intelligence*, 2017.

---
[1]UAN [7]: https://github.com/thuml/Universal-Domain-Adaptation

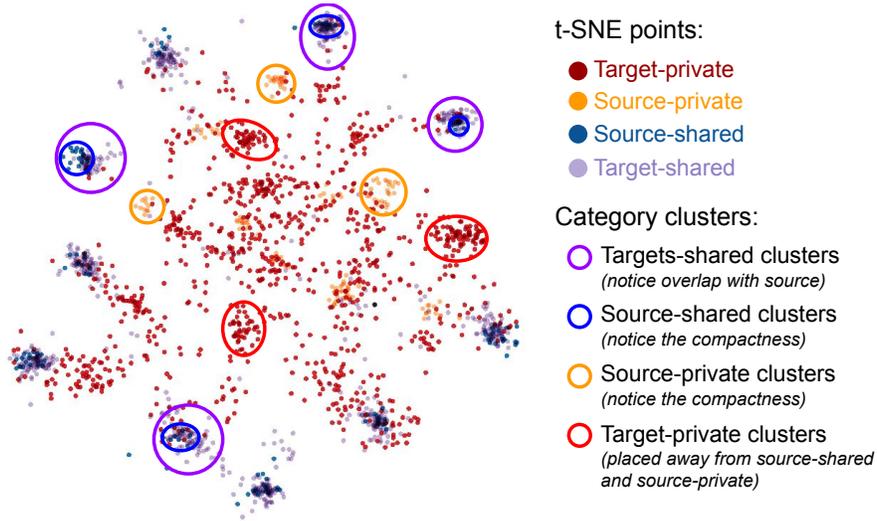

Figure 3: t-SNE plot showing the placement of all the four clusters computed after adaptation for the task A→D in Office-31. It validates our hypothesis in both the *Procurement* and the *Deployment* stages as shown by the highlighted clusters and the corresponding inferences in the legend under "Category clusters".

Table 5: Network architecture for the *Procurement* stage. Hyperparameter $\alpha = 0.2$

| Component | Trainable? | Operation | Notation | Features | Batch Norm? | Non-Linearity |
|---|---|---|---|---|---|---|
| **Resnet-50** (Upto AvgPool layer) | ✗ | | $M$ | 2048 | | |
| **Feature Extractor** | ✓ | | $F_s$ | 256 | | |
| | | Input | | 2048 | ✗ | |
| | | Fully connected | | 1024 | ✗ | ELU |
| | | Fully connected | | 1024 | ✓ | ELU |
| | | Fully connected | | 256 | ✗ | ELU |
| | | Fully connected | | 256 | ✓ | ELU |
| **Decoder** | ✓ | | $G$ | 2048 | | |
| | | Input | | 256 | ✗ | |
| | | Fully connected | | 1024 | ✗ | ELU |
| | | Fully connected | | 1024 | ✓ | ELU |
| | | Fully connected | | 2048 | ✗ | ELU |
| | | Fully connected | | 2048 | ✗ | - |
| **Classifier** | ✓ | | $D$ | $|\mathcal{C}_s| + |\mathcal{C}_n|$ | | |
| | | Input | | 256 | ✗ | |
| | | Fully connected | $D_{src}$ | $|\mathcal{C}_s|$ | ✗ | |
| | | Input | | 256 | ✗ | |
| | | Fully connected | $D_{neg}$ | $|\mathcal{C}_n|$ | ✗ | |

Table 6: Network architecture for the *Deployment* stage. Hyperparameter $\beta = 0.1$

| Component | Trainable? | Operation | Notation | Features | Batch Norm? | Non-Linearity |
|---|---|---|---|---|---|---|
| **Resnet-50** (Upto AvgPool layer) | ✗ | | $M$ | 2048 | | |
| **Feature Extractor** | ✓ | | $F_t$ | 256 | | |
| | | Input | | 2048 | ✗ | |
| | | Fully connected | | 1024 | ✗ | ELU |
| | | Fully connected | | 1024 | ✓ | ELU |
| | | Fully connected | | 256 | ✗ | ELU |
| | | Fully connected | | 256 | ✓ | ELU |
| **Classifier** | ✗ | | $D$ | $|\mathcal{C}_s| + |\mathcal{C}_n|$ | | |
| | | Input | | 256 | ✗ | |
| | | Fully connected | $D_{src}$ | $|\mathcal{C}_s|$ | ✗ | |
| | | Input | | 256 | ✗ | |
| | | Fully connected | $D_{neg}$ | $|\mathcal{C}_n|$ | ✗ | |

# References

[1] Mahsa Baktashmotlagh, Masoud Faraki, Tom Drummond, and Mathieu Salzmann. Learning factorized representations for open-set domain adaptation. In *ICLR*, 2019. 1

[2] Shai Ben-David, John Blitzer, Koby Crammer, and Fernando Pereira. Analysis of representations for domain adaptation. In *NeurIPS*, 2007. 1

[3] Konstantinos Bousmalis, Nathan Silberman, David Dohan, Dumitru Erhan, and Dilip Krishnan. Unsupervised pixel-level domain adaptation with generative adversarial networks. In *CVPR*, 2017. 2, 3

[4] P. P. Busto and J. Gall. Open set domain adaptation. In *ICCV*, 2017. 2

[5] Zhangjie Cao, Mingsheng Long, Jianmin Wang, and Michael I Jordan. Partial transfer learning with selective adversarial networks. In *CVPR*, 2018. 1, 2, 5

[6] Zhangjie Cao, Lijia Ma, Mingsheng Long, and Jianmin Wang. Partial adversarial domain adaptation. In *ECCV*, 2018. 1, 2, 5

[7] Yi-Hsin Chen, Wei-Yu Chen, Yu-Ting Chen, Bo-Cheng Tsai, Yu-Chiang Frank Wang, and Min Sun. No more discrimination: Cross city adaptation of road scene segmenters. In *ICCV*, 2017. 1

[8] Lixin Duan, Ivor W Tsang, and Dong Xu. Domain transfer multiple kernel learning. *TPAMI*, 34(3):465–479, 2012. 2

[9] Yaroslav Ganin, Evgeniya Ustinova, Hana Ajakan, Pascal Germain, Hugo Larochelle, François Laviolette, Mario Marchand, and Victor Lempitsky. Domain-adversarial training of neural networks. *The Journal of Machine Learning Research*, 17(1):2096–2030, 2016. 1

[10] ZongYuan Ge, Sergey Demyanov, Zetao Chen, and Rahil Garnavi. Generative openmax for multi-class open set classification. In *BMVC*, 2017. 1

[11] Ian Goodfellow, Jean Pouget-Abadie, Mehdi Mirza, Bing Xu, David Warde-Farley, Sherjil Ozair, Aaron Courville, and Yoshua Bengio. Generative adversarial nets. In *NeurIPS*, 2014. 2

[12] Yves Grandvalet and Yoshua Bengio. Semi-supervised learning by entropy minimization. In *NeurIPS*, 2005. 5

[13] Kaiming He, Xiangyu Zhang, Shaoqing Ren, and Jian Sun. Deep residual learning for image recognition. In *CVPR*, 2016. 6, 7, 8

[14] Dan Hendrycks, Mantas Mazeika, and Thomas Dietterich. Deep anomaly detection with outlier exposure. In *ICLR*, 2019. 2

[15] Judy Hoffman, Eric Tzeng, Taesung Park, Jun-Yan Zhu, Phillip Isola, Kate Saenko, Alexei A Efros, and Trevor Darrell. Cycada: Cycle-consistent adversarial domain adaptation. In *ICLR*, 2018. 2

[16] Lanqing Hu, Meina Kan, Shiguang Shan, and Xilin Chen. Duplex generative adversarial network for unsupervised domain adaptation. In *CVPR*, 2018. 2

[17] Guoliang Kang, Liang Zheng, Yan Yan, and Yi Yang. Deep adversarial attention alignment for unsupervised domain adaptation: the benefit of target expectation maximization. In *ECCV*, 2018. 2

[18] Diederik P Kingma and Jimmy Ba. Adam: A method for stochastic optimization. *arXiv preprint arXiv:1412.6980*, 2014. 6

[19] Diederik P Kingma and Max Welling. Auto-encoding variational bayes. *arXiv preprint arXiv:1312.6114*, 2013. 4

[20] Alex Krizhevsky, Ilya Sutskever, and Geoffrey E Hinton. Imagenet classification with deep convolutional neural networks. In *NeurIPS*, 2012. 1

[21] Abhishek Kumar, Prasanna Sattigeri, Kahini Wadhawan, Leonid Karlinsky, Rogerio Feris, Bill Freeman, and Gregory Wornell. Co-regularized alignment for unsupervised domain adaptation. In *NeurIPS*, 2018. 1

[22] Jogendra Nath Kundu, Nishank Lakkakula, and R Venkatesh Babu. Um-adapt: Unsupervised multi-task adaptation using adversarial cross-task distillation. In *ICCV*, 2019. 1

[23] Christoph H Lampert, Hannes Nickisch, and Stefan Harmeling. Learning to detect unseen object classes by between-class attribute transfer. In *CVPR*, 2009. 4

[24] Kimin Lee, Honglak Lee, Kibok Lee, and Jinwoo Shin. Training confidence-calibrated classifiers for detecting out-of-distribution samples. In *ICLR*, 2018. 2

[25] Zhizhong Li and Derek Hoiem. Learning without forgetting. *TPAMI*, 40(12):2935–2947, 2017. 2

[26] Mingsheng Long, Yue Cao, Jianmin Wang, and Michael Jordan. Learning transferable features with deep adaptation networks. In *ICML*, 2015. 1, 2

[27] Mingsheng Long, Zhangjie Cao, Jianmin Wang, and Michael I Jordan. Conditional adversarial domain adaptation. In *NeurIPS*, 2018. 2

[28] Mingsheng Long, Han Zhu, Jianmin Wang, and Michael I Jordan. Unsupervised domain adaptation with residual transfer networks. In *NeurIPS*, 2016. 1, 2, 5

[29] Raphael Gontijo Lopes, Stefano Fenu, and Thad Starner. Data-free knowledge distillation for deep neural networks. In *LLD Workshop at NeurIPS*, 2017. 2

[30] Zelun Luo, Yuliang Zou, Judy Hoffman, and Li F Fei-Fei. Label efficient learning of transferable representations acrosss domains and tasks. In *NeurIPS*, 2017. 2

[31] Andrey Malinin and Mark Gales. Predictive uncertainty estimation via prior networks. In *NeurIPS*, 2018. 1

[32] Jogendra Nath Kundu, Phani Krishna Uppala, Anuj Pahuja, and R Venkatesh Babu. Adadepth: Unsupervised content congruent adaptation for depth estimation. In *CVPR*, 2018. 1, 2

[33] Pau Panareda Busto and Juergen Gall. Open set domain adaptation. In *ICCV*, 2017. 2, 7, 8

[34] Xingchao Peng, Ben Usman, Neela Kaushik, Judy Hoffman, Dequan Wang, and Kate Saenko. Visda: The visual domain adaptation challenge. In *CVPR workshops*, 2018. 6

[35] Shaoqing Ren, Kaiming He, Ross Girshick, and Jian Sun. Faster r-cnn: Towards real-time object detection with region proposal networks. In *NeurIPS*, 2015. 1

[36] Olga Russakovsky, Jia Deng, Hao Su, Jonathan Krause, Sanjeev Satheesh, Sean Ma, Zhiheng Huang, Andrej Karpathy, Aditya Khosla, Michael Bernstein, et al. Imagenet large scale visual recognition challenge. *IJCV*, 115(3):211–252, 2015. 2, 3, 6

[37] Kate Saenko, Brian Kulis, Mario Fritz, and Trevor Darrell. Adapting visual category models to new domains. In *ECCV*, 2010. 2, 6

[38] Kuniaki Saito, Kohei Watanabe, Yoshitaka Ushiku, and Tatsuya Harada. Maximum classifier discrepancy for unsupervised domain adaptation. In *CVPR*, 2018. 1, 2, 3, 5

[39] Kuniaki Saito, Shohei Yamamoto, Yoshitaka Ushiku, and Tatsuya Harada. Open set domain adaptation by backpropagation. In *ECCV*, 2018. 1, 2, 6, 7, 8

[40] Tim Salimans, Ian Goodfellow, Wojciech Zaremba, Vicki Cheung, Alec Radford, and Xi Chen. Improved techniques for training gans. In *NeurIPS*, 2016. 2, 3

[41] Swami Sankaranarayanan, Yogesh Balaji, Carlos D Castillo, and Rama Chellappa. Generate to adapt: Aligning domains using generative adversarial networks. In *CVPR*, 2018. 2, 3

[42] Alireza Shafaei, Mark Schmidt, and James Little. A Less Biased Evaluation of Out-of-distribution Sample Detectors. In *BMVC*, 2019. 2

[43] Hidetoshi Shimodaira. Improving predictive inference under covariate shift by weighting the log-likelihood function. *Journal of statistical planning and inference*, 90(2):227–244, 2000. 1

[44] Rui Shu, Hung Bui, Hirokazu Narui, and Stefano Ermon. A DIRT-t approach to unsupervised domain adaptation. In *ICLR*, 2018. 4

[45] Eric Tzeng, Judy Hoffman, Trevor Darrell, and Kate Saenko. Simultaneous deep transfer across domains and tasks. In *ICCV*, 2015. 1

[46] Eric Tzeng, Judy Hoffman, Kate Saenko, and Trevor Darrell. Adversarial discriminative domain adaptation. In *CVPR*, 2017. 1, 2, 3, 5

[47] Hemanth Venkateswara, Jose Eusebio, Shayok Chakraborty, and Sethuraman Panchanathan. Deep hashing network for unsupervised domain adaptation. In *CVPR*, 2017. 6

[48] Oriol Vinyals, Charles Blundell, Timothy Lillicrap, Daan Wierstra, et al. Matching networks for one shot learning. In *NeurIPS*, 2016. 4

[49] Xuezhi Wang and Jeff Schneider. Flexible transfer learning under support and model shift. In *NeurIPS*, 2014. 2

[50] Zirui Wang, Zihang Dai, Barnabás Póczos, and Jaime Carbonell. Characterizing and avoiding negative transfer. In *CVPR*, 2019. 2

[51] Ancong Wu, Wei-Shi Zheng, Xiaowei Guo, and Jian-Huang Lai. Distilled person re-identification: Towards a more scalable system. In *CVPR*, 2019. 2

[52] Kaichao You, Mingsheng Long, Zhangjie Cao, Jianmin Wang, and Michael I. Jordan. Universal domain adaptation. In *CVPR*, June 2019. 3, 5, 6, 7, 8

[53] Hongyi Zhang, Moustapha Cisse, Yann N. Dauphin, and David Lopez-Paz. mixup: Beyond empirical risk minimization. In *ICLR*, 2018. 4

[54] Jing Zhang, Zewei Ding, Wanqing Li, and Philip Ogunbona. Importance weighted adversarial nets for partial domain adaptation. In *CVPR*, 2018. 1, 2, 7, 8

[55] Kun Zhang, Bernhard Schölkopf, Krikamol Muandet, and Zhikun Wang. Domain adaptation under target and conditional shift. In *ICML*, 2013. 2

[56] Weichen Zhang, Wanli Ouyang, Wen Li, and Dong Xu. Collaborative and adversarial network for unsupervised domain adaptation. In *CVPR*, 2018. 1, 2

[57] Jun-Yan Zhu, Taesung Park, Phillip Isola, and Alexei A Efros. Unpaired image-to-image translation using cycle-consistent adversarial networks. In *ICCV*, 2017. 2



\end{document}